\documentclass[preprint,12pt]{elsarticle}
\usepackage{color}

\usepackage{amssymb}
\usepackage{times}  
\usepackage{helvet}  
\usepackage{courier}  
\usepackage[hyphens]{url}  
\usepackage{graphicx} 
\usepackage{natbib}  
\usepackage{caption} 
\usepackage{amsmath}
\DeclareMathOperator{\EX}{\mathbb{E}}
\usepackage{algorithm}
\usepackage{algorithmic}
\usepackage{booktabs}
\usepackage{xcolor} 

\journal{International Journal of Multimedia Information Retrieval}

\begin{document}

\begin{frontmatter}



\title{ANROT-HELANet: Adverserially and Naturally Robust Attention-Based Aggregation Network via The Hellinger Distance for Few-Shot Classification}


\author[label1,label4]{Gao Yu Lee}

\author[label2]{Tanmoy Dam}

\author[label3]{Md Meftahul Ferdaus}

\author[label1]{Daniel Puiu Poenar}

\author[label4]{Vu N.Duong}

\affiliation[label1]{organization={School of Electrical and Electronic Engineering, Nanyang Technological University},
            addressline={50 Nanyang avenue}, 
            city={Singapore},
            postcode={639798}, 
            country={Singapore}}

\affiliation[label2]{organization={SAAB-NTU Joint Lab, School of Mechanical and Aerospace Engineering, Nanyang Technological University},
            addressline={50 Nanyang avenue}, 
            city={Singapore},
            postcode={639798}, 
            country={Singapore}}

\affiliation[label3]{organization={Department of Computer Science, The University of New Orleans},
            addressline={2000 Lakeshore Drive}, 
            city={New Orleans},
            postcode={70148}, 
            state={LA},
            country={United States}}
            
\affiliation[label4]{organization={Air Traffic Management Research Institute, School of Mechanical and Aerospace Engineering, Nanyang Technological University},
            addressline={50 Nanyang avenue}, 
            city={Singapore},
            postcode={639798}, 
            country={Singapore}}

\begin{abstract}

Few-Shot Learning (FSL), which involves learning to generalize using only a few data samples, has demonstrated promising and superior performances to ordinary CNN methods. While Bayesian based estimation approaches using Kullback-Leibler (KL) divergence have shown improvements, they remain vulnerable to adversarial attacks and natural noises. We introduce ANROT-HELANet, an Adversarially and Naturally RObusT Hellinger Aggregation Network that significantly advances the state-of-the-art in FSL robustness and performance. Our approach implements an adversarially and naturally robust Hellinger distance-based feature class aggregation scheme, demonstrating resilience to adversarial perturbations up to $\epsilon=0.30$ and Gaussian noise up to $\sigma=0.30$. The network achieves substantial improvements across benchmark datasets, including gains of 1.20\% and 1.40\% for 1-shot and 5-shot scenarios on miniImageNet respectively. We introduce a novel Hellinger Similarity contrastive loss function that generalizes cosine similarity contrastive loss for variational few-shot inference scenarios. Our approach also achieves superior image reconstruction quality with a FID score of 2.75, outperforming traditional VAE (3.43) and WAE (3.38) approaches. Extensive experiments conducted on four few-shot benchmarked datasets verify that ANROT-HELANet's combination of Hellinger distance-based feature aggregation, attention mechanisms, and our novel loss function establishes new state-of-the-art performance while maintaining robustness against both adversarial and natural perturbations. 
\end{abstract}




\end{frontmatter}


\begin{table}[h!]
\centering
\renewcommand{\arraystretch}{1.0}
\caption{List of acronyms and their full forms}
\label{tab:acronyms}
{\begin{tabular}{lp{0.65\textwidth}}
\toprule
\textbf{Acronym} & \textbf{Definition} \\
\midrule
ANROT-HELANet & Adversarially and Naturally Robust Hellinger Aggregation Network \\
ARFSIC & Adversarially Robust Few-Shot Image Classification \\
BC & Bhattacharyya Coefficient \\
CIFAR-FS & CIFAR-Few Shot (a subset of CIFAR-100) \\
CNN & Convolutional Neural Network \\
$D_{B}$ & Bhattacharyya Distance \\
$D_{H}$ & Hellinger Distance \\
$D_{KL}$ & Kullback-Leibler Divergence \\
$D_{M}$ & Mahalanobis Distance \\
$D_{W}$ & Wasserstein Distance \\
ELBO & Evidence Lower Bound \\
FC-100 & Few-shot Classification subset of CIFAR-100 \\
FGSM & Fast Gradient Sign Method \\
FID & Fréchet Inception Distance \\
FSL & Few-Shot Learning \\
GPU & Graphics Processing Unit \\
GRAD-CAM & Gradient-weighted Class Activation Mapping \\
HeSim & Hellinger Similarity \\
JoCoR & Joint Training with Co-Regularization \\
KL & Kullback–Leibler \\
$\mathcal{L}_{CCE}$ & Categorical Cross Entropy loss \\
$\mathcal{L}_{Hesim}$ & Hellinger Similarity loss function \\
$\mathcal{L}_{rec}$ & Reconstruction loss function \\
miniImageNet & A commonly used 100-class subset of ImageNet for few-shot learning \\
MLP & Multi-Layer Perceptron \\
RNNP & Robust Nearest Neighbor Prototype-based testing \\
SOTA & State-Of-The-Art \\
tieredImageNet & A larger subset of ImageNet with 608 classes for few-shot learning \\
VAE & Variational AutoEncoder \\
WAE & Wasserstein AutoEncoder \\
\bottomrule
\end{tabular}}
\end{table}

\section{Introduction}
Computer vision has made tremendous strides over the recent decades, many of which may be ascribed to deep supervised learning as well as the accessibility of vast amounts of data. This, however, still falls well short of our human visual and brain system's capacity to learn, recognize and adapt to new objects and classes despite exposure to only limited samples. Few-Shot Learning (FSL, see Table 1 for a complete list of acronyms) has recently been proposed as a viable solution to imitate the aforementioned recognition capabilities. FSL has demonstrated promising results relative to conventional CNN-based methods, and thus research interests in this area have rapidly increased, especially on situation in which the AI systems are continuously fed with new data and classes and be required to learn them on-the-go. An example of a recent review of FSL (in the context of remote sensing) can be found at Lee et al. \cite{lee2024unlocking}. 

FSL methods typically deployed a $N$-way-$k$-shot meta-training and evaluation strategy, where $N$ is the dataset's number of classes and $k$ is the number of samples in each class. A ``\emph {support}" and ``\emph{query}" set are also involved, where the former is used for training and the latter is used for prediction. Prior FSL approaches determines the degree of similarity between images and establishing a distance measure assigning the same class label to similarly shaped objects and a different class label to wildly dissimilar objects. All these can be carried out either directly (e.g., Siamese network; see Koch et al. \cite{koch2015siamese}), or by processing the visual information into a higher-dimensional embedding space; see prototypical network by Snell et al. \cite{snell2017prototypical}. For the latter, this leads to embedded feature points of the same class clustering in the same region. The prototypical network specifically utilized a feature point estimate technique, which despite being demonstrated to produce State-Of-the-Art (SOTA) performance in several benchmark datasets, nevertheless has some drawbacks, as pointed out by Zhang et al. \cite{zhang2019variational}. This includes 

\begin{itemize}
    \item erroneous point estimations if there are only few unevenly distributed support sample points,
    \item insufficient interpretability for class designation when using only a single embedding of the feature point,
    \item overfitting if there is any discrepancy in the data distribution between the training and test phases.
\end{itemize}

 To overcome the above limitations, a probabilistic distribution-based estimation strategy was proposed by Zhang et al. \cite{zhang2019variational}, which utilized a Bayesian-based estimation approach that is useful in data scarcity scenario, hence emphasizing the essential role that contextual prior knowledges can play. However, an intractable integral must be computed. Therefore, a variational approximation was used, which minimizes the Kullback-Leibler (KL) divergence, and thus maximizing an Evidence Lower Bound (ELBO). The probability distributions to be compared include $p(z|\mathcal{T})$ and a posterior distribution $q(z|\mathcal{S})$, where $\mathcal{S}$ refers to the query set and $\mathcal{T}$ is the support set. Therefore, the KL divergence takes on the form $D_{KL}(q(z|\mathcal{S})|| p(z|\mathcal{T}))$. The predicted labels after training, rather than being obtained directly, are obtained via computing the confidence scores for a given meta-task involving $Q$ number of query samples. The KL divergence is however, asymmetrical, and hence any swapping of the two distributions throughout a model's process could produce different findings, making any comparisons challenging. The Hellinger distance \cite{hellinger1909neue} is a viable alternative to the KL divergence in that not only it is symmetrical, it is the probabilistic equivalent of the Euclidean distance, thus making computation and comparison simpler. Furthermore, to the best of our knowledge, there are existing works on utilizing the Hellinger distance for classification in other domains (e.g., \cite{grzyb2021hellinger}, \cite{kumari2017hellinger}), while little attention have been focused on incorporating such distance in FSL classification.

\begin{figure}[hbt!]
    \centering
    \includegraphics[scale=0.75]{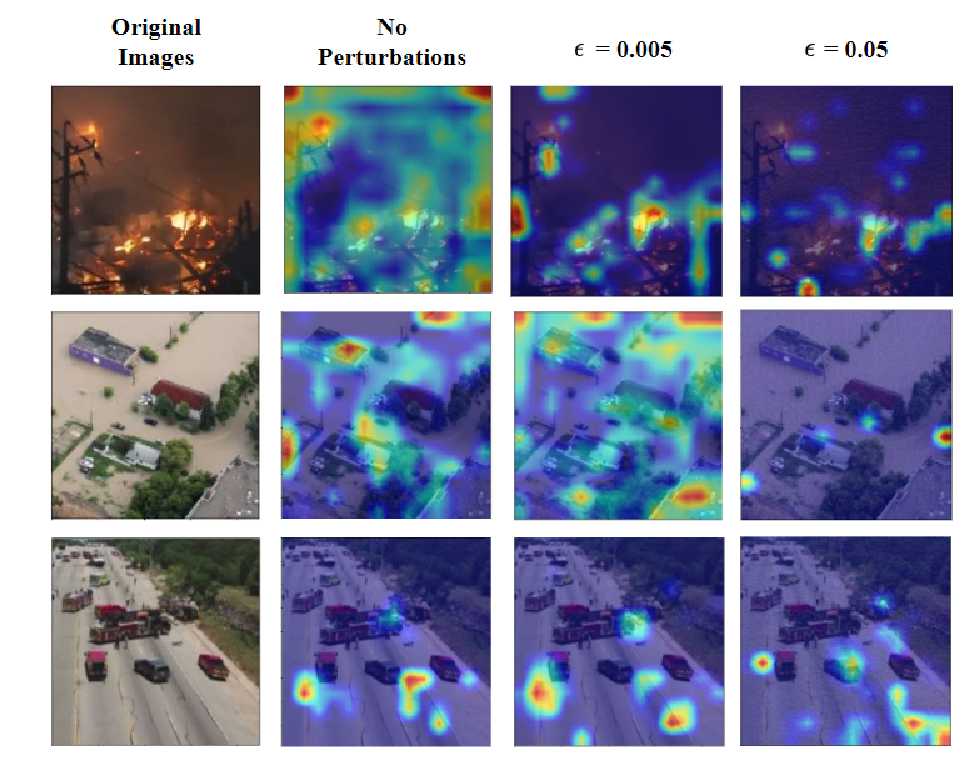}
    \caption{Some illustrations of how even a small adversarial perturbations ($\epsilon = 0.05$) can drastically affect the attention heatmap captured using GRAD-CAM. For the fire images (top), we can see that by increasing the perturbation factor $\epsilon$, the heatmap gradually shift from the essential fire hotspots and smokes to non-fire spots. Such shifting is also observed for the flood (middle) and traffic accident images (bottom). The images are adapted from Kyrkou and Theocharides \cite{kyrkou2020emergencynet}.}
    \label{fig:AIDER_fgsm_figures}
\end{figure}

In addition, adversarial training-based techniques has gradually being applied in few-shot learning. The motivation behind this direction is that current SOTAs few-shot approaches are still highly vulnerable to adversarial examples and attacks, as emphasized by Goldblum et al. \cite{goldblum2020adversarially} and Dong et al. \cite{dong2022improving}. The role of the adversarial training scheme is to not only generate adversarial samples, but to also make the model more adversarially robust along with fine-tuning during training. Such models are also known as Adversarially Robust Few-Shot Image Classification (ARFSIC) model. The importance of such models lies in situations involving scarce data, or the need to constantly adapt to new classes, all while simultaneously resisting the attacks that can lead to catastrophic consequences. (e.g., in autonomous driving). Initial adversarial robust models does not incorporate attention mechanism, which could potentially enhance the models in a more efficient manner via focusing only on the key features and disregard the less important ones, as well as allowing a visualization of the image's regions that are responsible for the model's decision in any classification or mis-classification \cite{wu2018attention}. Following Wang et al. \cite{wang2021agkd}, we illustrate in Figure \ref{fig:AIDER_fgsm_figures} how bringing in attention could help illustrate how the attacks leads to the shift in the heatmap from relevant features pertinent in a given object of certain classes to the irrelevant features. Furthermore, Agrawal et al. \cite{agrawal2021impact} and Hsieh et al. \cite{hsieh2019robustness} demonstrated that attention-based adversarial models are more robust than the non-attention ones (although it may be dependent on the number of classes in the dataset). The different types of attention-based approaches that has been incorporated for the ARFSIC include self-attention \cite{mu2021defending}, attention-transfer \cite{li2021adversarial}, and combining attention with Logit Pairing \cite{goodman2019improving}. However, as of current, there are little works that explored the role of the attention mechanism in ARFSIC to enhance the model's robustness. Such issue is even more of a pressing need for the reasons described previously. Furthermore, it can shed light into how the model attends to essential features under scarce data resources per training phase.

Lastly, apart from adversarial noises, noises that are produced by natural sources such as changes in weather conditions, sensor capability changes, presence of hazes or rains, and changes in object features with time, can be detrimental to neural network performances. These changes can constitute a natural shift of the feature distribution and hence can easily mislead a target class or object to be identified. Liu et al. \cite{liu2023comprehensive} provided a succinct summary of this issue, and has summarized and proposed various benchmark datasets to assess the impact of a proposed algorithm's natural robustness. These datasets include ObjectNet \cite{barbu2019objectnet} which involved random changes to the image backgrounds, rotations and viewpoints, ImageNet-R \cite{hendrycks2021many}, which involved changes in image style, and ImageNet-C \cite{hendrycks2019benchmarking} which involved synthesized corruption of five levels of severity such as those due to image blurring and weather conditions. However, these dataset are more towards testing the natural robustness of ordinary CNN-based training and inference. As far as we know (until very recently \cite{mazumder2021rnnp}), there are little works that focus on few-shot-based naturally robust approaches, as well as the respective benchmark datasets that served as a platform for the robustness assessment. Hence, this leaves the question of how incorporating adversarial or natural noise samples in the few-shot data domain in images could help in improving robustness performance of the proposed network.

Therefore, we introduced the Adversarially and Naturally RObusT-HELlinger Aggregation Network (ANROT-HELANet) which performs an ARFSIC and naturally robust training paradigm via the Hellinger distance. The upshot of our approach is that instead of the usual Kullback-Leibler divergence, the Hellinger distance is utilized instead. Since both measures are a subset of the general \emph{f-divergences}, this means that optimization principles that are utilized in KL divergences based methods can be easily be generalize to our selected distance metric. This can be done due to our capability to derive and compute an ELBO in terms of the Hellinger distance and the evidence term. Furthermore, we introduced an Hellinger Similarity loss function ($\mathcal{L}_{Hesim}$) which is inspired from the cosine similarity loss implemented in Chen et al. \cite{chen2020simple}, but replaces the cosine similarity in the numerator with the Hellinger similarity. Therefore, apart from being adversarially and naturally robust, our network has been shown to yield superior classification accuracy values in the established few-shot evaluation approach. 

Furthermore, we provided theoretical verification of $\mathcal{L}_{Hesim}$ performance in high-dimensional space and its specific impact on few-shot learning tasks. Since the Hellinger distance can be associated with the Mahalanobis distance $\mathcal{D}_M$ via the Bhattarcharyya coefficient $BC$ according to equation (25), and given that the Mahalanobis distance effectively captures data correlations in high-dimensional space while accounting for the scale of different features, our $\mathcal{L}_{Hesim}$ inherits these beneficial properties for high-dimensional feature representation. This theoretical foundation is particularly crucial in few-shot learning scenarios where the model must learn effective feature representations from limited samples. The impact of $\mathcal{L}_{Hesim}$ on few-shot learning tasks is manifested through its ability to better capture the probabilistic nature of class prototypes in the embedding space, as demonstrated by the superior classification accuracies achieved across all benchmarked datasets (detailed in Section 5.1). Additionally, our ablation studies (Section 5.4) empirically verify that configurations utilizing $\mathcal{L}_{Hesim}$ consistently outperform those using alternative loss functions, providing both theoretical and empirical validation of its effectiveness in few-shot learning scenarios.

In summary, our contributions are as follows:

\begin{itemize}
    \item Our proposed ANROT-HELANet pioneered the usage of the Hellinger distance to compute and cluster the data features into respective class prototypes via variational inferences. The adversarial robustness is achieved via generating the adversarial samples and performing fine-tuning.
    \item We introduced a Hellinger Similarity loss function $\mathcal{L}_{Hesim}$ which, combined with the reconstruction loss and the categorical cross-entropy loss, allow us to arrive at comparatively high-quality reconstructed images while maintaining a competitive FSL performances relative to the State-Of-The-Art (SOTA).
    \item As far as we know, we are one of the first work that takes into account both adversarial and natural noise perturbations in the few-shot training paradigm, and utilize both adversarial and natural robustness training to further enhance our network's resistance to such undesired effects.
    \item Extensive experiments on four commonly-benchmarked few-shot image dataset that had been perturbed with adversarial and Gaussian natural noise, which includes CIFAR-100 \cite{krizhevsky2009learning}, FC-100 \cite{bertinetto2018meta}, miniImageNet \cite{vinyals2016matching} and tieredImageNet \cite{ren2018meta} demonstrated the feasibility and superior accuracy performances relative to the SOTAs. Additionally, we performed a reconstruction image quality evaluation of the Hellinger distance in our model relative to the KL divergence and the Wasserstein distance, and demonstrated that the Hellinger distance yielded better quality outputs than the other two. Hence our ANROT-HELANet is preferential in generating novel and higher-resolution image to enhance few-shot training.
\end{itemize}

The theoretical advantages of simultaneously incorporating both adversarial and natural noise perturbations in few-shot learning can be understood through regularization theory and robust feature learning principles. In few-shot scenarios where limited training samples are available, the model is particularly susceptible to overfitting to incorrect features. When combined, these perturbations create complementary regularization effects: adversarial training enforces local Lipschitz continuity in the model's decision boundary around training points by targeting the model's most sensitive directions in feature space, while natural noise training simulates real-world distribution shifts the model may encounter during deployment.

The dual regularization proves effective because adversarial perturbations force the model to focus on robust features that remain discriminative under worst-case scenarios, while natural noise augmentation expands the training distribution to reflect real-world variations. This combination creates a more comprehensive robustness framework. Adversarial training alone results in fragile robustness against specific perturbation patterns, and natural noise training alone does not capture adversarial vulnerability patterns. However, their integration enables the model to develop both targeted defense mechanisms and broad-spectrum resilience. Our model maintained classification accuracy under adversarial perturbations up to $\epsilon=0.30$ while demonstrating robustness to Gaussian noise perturbations up to $\sigma=0.30$.

\section{Related Works}

In this section we provide some background information and existing work done in the relevant area of adversarial robustness and few-shot variational learning. 

\subsection{Few-Shot Variational Learning}

Apart from \cite{zhang2019variational} as described earlier, Han et al. \cite{han2023few} proposed a variational few-shot feature aggregation strategy that merged the support and query features during batch training to enhance object detection. This fosters inter-class interactions, which in turn encourages representations that are not class-specific, lessening the confusion in the prediction phase between the base and novel classes. The transductive decoupled variational inference network (TRIDENT) was introduced by Singh and Jamali-Rad \cite{singh2022transductive}, which performed inference by fusing the informative in the support and query set, akin to the feature aggregation work, via decoupling the image representation into semantic and label latent variables. The TRIDENT included attention during the feature extraction stage unlike the two approaches stated above. We also discovered that additional attention-based FSL works exist, such as Ngyuen et al. \cite{nguyen2022enhancing}, but such works are more focused on query point embedding. As far as we know, TRIDENT is the first FSL study to combine attention and variational inference, and has demonstrated its feasibility to improve classification performances. All of the approaches discussed, however, approached the log marginal likelihood optimization via the association between the ELBO and KL divergence. Our proposed method breaks this convention by deriving an association between the ELBO and Hellinger distance, thus providing a feasible and effective alternative to the domain of variational-based FSL methodology.

\subsection{Emerging Few-Shot Methods}
Recent developments in few-shot learning have seen significant advances in self-supervised learning approaches and adversarial training-based methods. Self-supervised few-shot learning methods aim to learn representations without relying on labeled data, offering advantages in scenarios with limited annotation. For instance, Chen et al.\cite{Chen2022} demonstrated strong performance through contrastive learning, though their approach can struggle with highly similar classes due to the reliance on instance discrimination. Similarly, while approaches like MELR \cite{Yan2024} effectively leverage meta-learning with self-supervised regularization, they may require substantial computational resources for pre-training.

In parallel, recent advances in adversarial training for few-shot scenarios have introduced various strategies beyond traditional Fast Gradient Sign Method (FGSM) \cite{goodfellow2014explaining}-based approaches. Methods such as FEAT \cite{Fu2023} combine adversarial training with self-attention mechanisms, showing improved robustness but potentially increased computational overhead. IEPT \cite{Dong2024} introduces instance-level pretext tasks for improved robustness, though its effectiveness can vary across different domains. Notably, while these methods have shown promise in handling adversarial attacks, they often focus less on natural robustness. Conversely, approaches specifically targeting natural robustness, such as Deep-EMD \cite{Nayak2022}, effectively handle distribution shifts but may not provide sufficient protection against targeted adversarial attacks.

The limitations of these emerging methods highlight the need for our integrated approach that simultaneously addresses both adversarial and natural robustness. While recent self-supervised methods provide powerful representation learning capabilities and adversarial training approaches offer targeted defense mechanisms, few existing approaches comprehensively address both aspects within a unified framework.

\subsection{Adversarial Robustness}

FGSM is one of the first adversarially sample generation approach, in which white-box attack-based robustness strategies had been put in place (e.g., Madry et al. \cite{madry2017towards}, and Xie et al.\cite{xie2019feature}). It was expected in the few-shot domain, a black-box attack-based robustness approach would be desirable, as no output labels are available \emph{a priori}. However, the approach by Goldblum et al. \cite{goldblum2020adversarially} and Dong et al. \cite{dong2022improving} utilized corresponding labels during the generation process, for which the corresponding samples were then inputted into the few-shot classifier. This is because the focus of these work is more on exploring the effect of including the adversarial samples during training on the classification performances, rather than assuming the black-box nature of the attack. Following this paradigm, our approach also considered the generation of the adversarial samples as a pre-processing step to our network training procedure. However, so far none of the method described incorporated the attention mechanism, either during the generation process, or in their respective adversarially robust loss function during training. Therefore, our approach aims to bridge this gap by also defining an attention-based adversarially robust loss function.

\subsection{Natural Robustness}

As described in the introductory section, there are little works that focus on analyzing and implementing naturally robust few-shot networks, to the best of our knowledge. However, one of the closer related works is the Robust Nearest Neighbor Prototype-based testing (RNNP) proposed by Mazumder et al. \cite{mazumder2021rnnp} which attempts to make the model robust to label corruption or noises presented in the support set. This is done via modifying the prototype computation stage, since the usual computation approach would also utilize the support set sample with the corrupt labels. This entail firstly the creation of unlabeled hybrid features via merging the support set features from the same class in a proportional manner, utilizing the hyperparameters $\alpha$ as a weighing factor (They have selected 0.8 for their experimental runs). Then for each query image, a soft k-means clustering is performed on the combined support image features, the hybridized features, and the query image features. Lastly, soft labels are assigned to the aforementioned features and is used to guide the cluster centres updating in an iterative manner to arrive at the refined prototypes. Such procedure helps reduce the influential effects of the corrupted labels on the class prototypes, and also reduces overfitting since the refined prototypes are arrived at using both the hybrid and the support features. The benchmark dataset utilized for their model is the miniImageNet and tieredImageNet, which is also part of our chosen dataset. Although improvements in the robustness performance were reported in the 5-shot and 10-shot setting, their model was only evaluated with respect to the Co-Teaching+ \cite{yu2019does} and the Joint Training with Co-Regularization (JoCoR) \cite{wei2020combating}, which are originally developed for the non few-shot regime. Also, their work focuses more on the scenario whereby the labels are corrupted, rather than on the impact of image corruption on the robustness. 

Our method address these gap in a three-fold manner: Firstly, our ANROT-HELANet aimed to assess its robustness performance in the more widely evaluated 1-shot and 5-shot setting, which allows for a fairer comparison of the values obtained from the state-of-the-art methods. This also means that more existing few-shot approaches can be selected for evaluation. Secondly, apart from the miniImageNet and tieredImageNet, we also utilized FC-100 and CIFAR-FS to expand the scope of our dataset benchmark to that generated by CIFAR-100 instead of ImageNet (Liu et al. \cite{liu2023comprehensive} listed out a list of natural robustness benchmark, for which all are variants of the ImageNet). This allows better generalization of our network to more few-shot classification classes and objects. Lastly, our work corrupt the image samples directly while keeping the labels intact, allowing us to explore yet another perspective of the impact of few-shot-based samples corruption on the robustness degradation. The next section will describe more on the type of adversarial and natural noise chosen for the corruption of the image samples.  

\section{Our Approach}

\subsection{Problem Setting}

In the exposition of our few-shot training paradigm, we posit the existence of a labeled training set designated as $X_{base} = \{x_{i}, y_{i}\}_{i=1}^{N_{base}}$, where $x_{i}$ represents the raw features pertaining to the $i^{th}$ sample, while $y_{i}$ signifies the one-hot encoded label associated with the same. This labeled ensemble is the meta-training (or base) dataset. Adhering to the FSL paradigm, we consider a meta-test dataset $X_{test} = \{x_{i}, y_{i}\}_{i=1}^{N_{test}}$ , comprising a novel set of classes represented by $Y_{test}$. The uniqueness of this set is encapsulated in the equation $Y_{base}\cap Y_{test} = \emptyset$, illustrating that $Y_{test}$ introduces completely new class entities not included in $Y_{base}$. The few-shot scenario involves the creation of randomly sampled few-shot tasks from this test dataset, each encompassing a limited number of labeled instances. Once training is done on the base set, FSL utilizes the labeled support sets for task-specific adaptation. The effectiveness of these techniques is then assessed based on their performance on the unlabeled query sets, providing a measure of the model's capacity to generalize to new data categories based on limited examples. 

\subsection{Adversarially and Naturally Robust Attention Mechanism}

\begin{figure}[hbt!]
    \centering
    \includegraphics[scale=0.55]{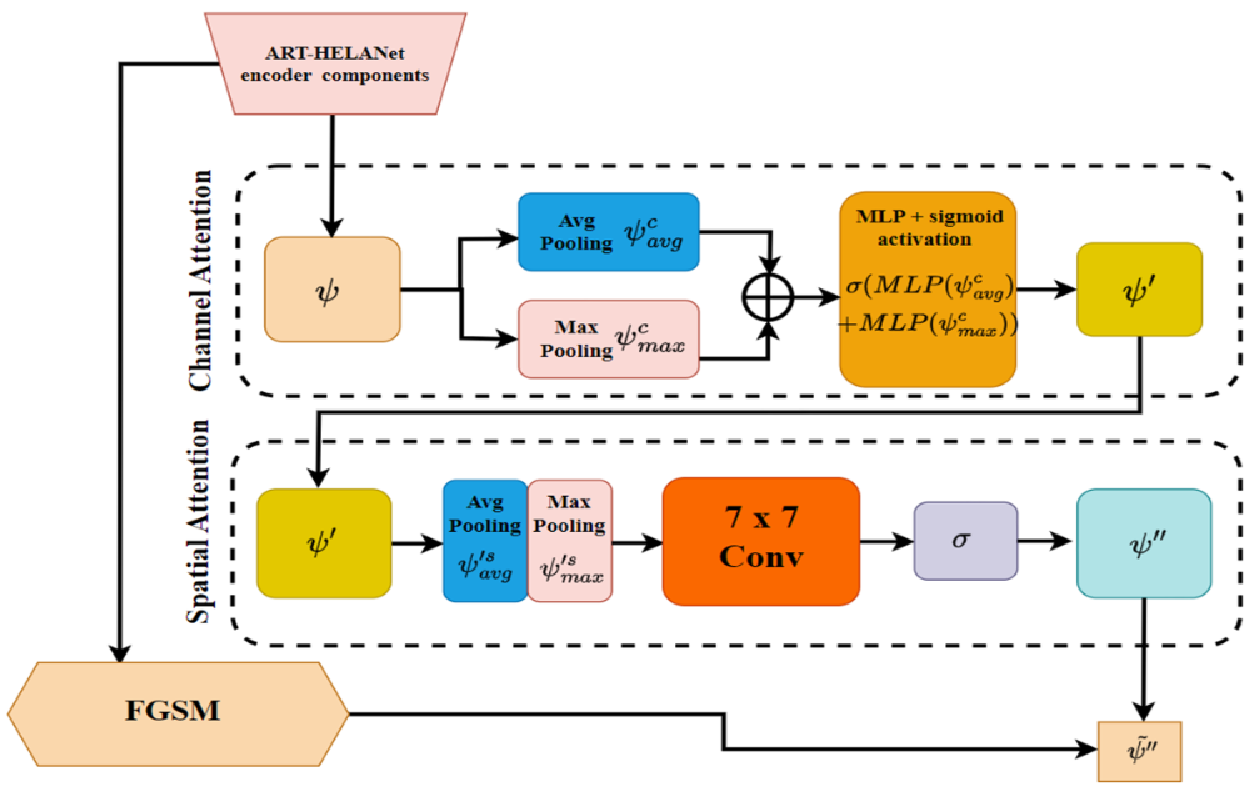}
    \caption{Algorithmic architecture of our attention mechanism along with the FGSM adversarial perturbations and the Gaussian natural noise training step.}
    \label{ART_HELANET}
\end{figure}

As mentioned, the highlight of our model is its capability to achieve both adversarial and natural robustness simultaneously. These two types of robustness address distinct challenges: artificial adversarial noise is deliberately crafted to deceive neural networks, while natural noise emerges from real-world conditions and environmental factors. Adversarial perturbations, such as those generated by FGSM \cite{goodfellow2014explaining}, are specifically designed to exploit model vulnerabilities by making minimal changes that cause misclassification. In contrast, natural noise encompasses real-world variations like changes in lighting conditions, weather effects, sensor noise, and compression artifacts that occur during image capture and processing.

We achieve ARFSIC by generating adversarial samples and performing adversarial training alongside the original training samples. This allows the model to adapt to discriminating patterns in the training data while simultaneously defending against adversarial examples generated via FGSM. Similarly, we achieve natural robustness by incorporating naturally corrupted samples during training. In our case, we selected Gaussian noise as it effectively approximates many common random noise patterns encountered in real-world image collection, such as sensor noise and thermal fluctuations. This dual approach ensures our model maintains robust performance both against potential adversarial attacks and under natural operating conditions where image quality may be degraded by environmental factors.

Unlike the original approach, we adopt an attention-based generation algorithm that generates both the adversarial and natural noise samples using a CNN architecture with the spatial and channel attention imbued at selected layers. The channel attention involved 4 operations: the max and average pooling, the Multi-Layer Perceptron ($MLP$), and a sigmoid activation function $\sigma$. The spatial attention involved 3 operations: the max and average pooling again, a convolutional layer with kernel size of 7 ($F^{7\times7}$), and the sigmoid activation function. The channel attention module processed the features inputs $\psi$ before the spatial attention module processed the corresponding outputs $\psi'$ from the channel attention. The spatial and channel attention mechanism is described mathematically in equation 1 and 2, with Figure \ref{ART_HELANET} outlining their algorithmic architecture.

\begin{equation} \label{eq1}
\begin{split}
\psi' = \mathcal{A}_{c}(\psi) &= \sigma(MLP(\psi^{c}_{avg}) + MLP(\psi^{c}_{max})),\\
\end{split}
\end{equation}

\begin{equation} \label{eq2}
\begin{split}
\psi'' = \mathcal{A}_{s}(\psi') &= \sigma(F^{7\times7}([\psi^{'s}_{avg} ;\psi^{'s}_{max}])). \\
\end{split}
\end{equation}

where $\mathcal{A}_{c}(\boldsymbol{\psi})$ and $\mathcal{A}_{s}(\psi')$ denotes the feature maps obtained from the channel and spatial attention respectively, $\psi^{c}_{avg}$ and $\psi^{c}_{max}$ represents the feature maps from the max and average pooling in the channel attention mechanism respectively, and $\psi^{'s}_{avg}$ and $\psi^{'s}_{max}$ are the feature maps from the max and average pooling in the spatial attention mechanism respectively. 

To attain adversarial robustness, FGSM is applied on $\psi''$, which is described mathematically as

\begin{equation} \label{eq3}
    \tilde{\psi''} = \psi'' + \epsilon \cdot sign(\boldsymbol{\nabla}_{\psi''(\boldsymbol{x})}(\psi''(\boldsymbol{w},\boldsymbol{x},\boldsymbol{y})),
\end{equation}

where $\psi''(\boldsymbol{x})$ are the feature maps associated with the inputs $\boldsymbol{x}$ from the attention mechanism, $\epsilon$ is the perturbation factor, $\boldsymbol{w}$ and $\boldsymbol{y}$ are the weight and ground-truth label, and $\tilde{\psi''}$ are the adversarially sampled feature maps which will form the adversarial images. To attain Gaussian noise robustness, instead of corrupting the extracted attention features $\psi$ directly, we corrupt the original images using Gaussian noises $\mathcal{N}(0, \sigma)$ of differing severity via experimenting with $\sigma$ values of 0.05, 0.10, 0.15, 0.20, 0.25 and 0.30, but keeping the mean values to 0. The training phase thus also involved a mix of the Gaussian noise corrupted samples and the original samples.

Finally, these images are fine-tuned during the few-shot training phase, as part of our ARFSIC and natural robustness scheme.

The selection of adversarial and natural noise parameters plays a crucial role in achieving optimal robustness while maintaining classification performance. For adversarial sample generation via FGSM, we carefully calibrated the perturbation factor $\epsilon$ within the range [0.05, 0.30]. This range was determined through extensive experimentation to balance meaningful adversarial perturbations with model stability. At $\epsilon=0.05$, our model maintains a high classification accuracy of 88.1\%, demonstrating strong resilience to small adversarial perturbations. The performance gradually degrades as $\epsilon$ increases, reaching 78.0\% at $\epsilon=0.10$, 73.9\% at $\epsilon=0.15$, and 68.6\% at $\epsilon=0.25$. This controlled degradation pattern indicates our model's ability to maintain reasonable performance even under stronger adversarial attacks.

For natural robustness, we implemented Gaussian noise perturbations with zero mean to preserve the data distribution center while varying the standard deviation $\sigma$ across [0.05, 0.30]. The selection of specific $\sigma$ values (0.05, 0.10, 0.15, 0.20, 0.25, and 0.30) provides a comprehensive evaluation of our model's robustness to natural variations. Our approach demonstrates remarkable stability under natural noise, with only a 3.7\% accuracy drop between $\sigma=0.15$ and $\sigma=0.20$, and maintaining 84.0\% accuracy even at $\sigma=0.30$. This robust performance can be attributed to several key factors in our architecture:

\begin{itemize} 
\item \textbf{Simultaneous Adversarial and Natural Robust Training.} Training with both adversarially perturbed and Gaussian-corrupted examples creates complementary regularization effects. The model thus learns targeted countermeasures for worst-case distortions (e.g., FGSM) as well as broader tolerance to naturally occurring corruptions. 
\item \textbf{Attention-Based Feature Extraction.} The spatial and channel attention mechanisms help the model focus on the most discriminative features, mitigating the impact of perturbations and leading to more stable performance across varied noise levels. 
\item \textbf{Hellinger Distance-Based Aggregation.} By leveraging the Hellinger distance (rather than relying solely on the KL divergence), our feature-aggregation step more effectively handles small distributional shifts. This confers additional stability in the presence of perturbations by capturing the underlying probabilistic relationships among noisy samples. 
\end{itemize}

Empirically, our parameter selection strategy, coupled with these architectural components, enables \emph{ANROT-HELANet} to achieve state-of-the-art classification robustness and accuracy across a broad spectrum of both adversarial and natural training conditions. Our experiments demonstrate that even as $\epsilon$ and $\sigma$ are increased, the network continues to exhibit graceful performance degradation while preserving much of its predictive power.

\subsection{ELBO and Hellinger Distance}

In our ANROT-HELANet, the goal is to minimize the divergence measure, which is equivalent to maximizing the ELBO. However the form of the ELBO bound and its association with the Hellinger distance ($D_{H}$) differs that of the KL divergence. We illustrate this starting with the definition of the square of $D_{H}$.



\begin{equation} \label{eq4}
\begin{split}
    D_{H}^{2} = 1 - \int\left(\sqrt{p_{\theta}(z|\mathcal{T})q_{\phi}(z|\mathcal{S})}\right) dz,
\end{split}
\end{equation}

where $p_{\theta}(z|\mathcal{T})$ described the prior distribution of $z$ with our knowledge of $\mathcal{T}$, and $q_{\phi}(z|\mathcal{S})$ denotes the true parameterized distribution conditioned on $\mathcal{S}$. Exchanging their positions and taking the logarithm of both sides yield, 

\begin{equation} \label{eq5}
    \textrm{log} \int\left(\sqrt{p_{\theta}(z|\mathcal{T})q_{\phi}(z|\mathcal{S})}\right) dz = \textrm{log} \left(1 - D^{2}_{H}\right).
\end{equation}

We note that the mathematical definition of the ELBO is

\begin{equation} \label{eq6}
    ELBO = \int q_{\phi}(z|\mathcal{S}) \textrm{log} \left(\frac{p_{\theta}(\mathcal{T},z)}{q_{\phi}(z|\mathcal{S})}\right) dz.
\end{equation}

Inserting the logarithm inside the integral in equation 5 on the left-hand side while bringing out the factor of $\frac{1}{2}$ and moving it to the term on the right-hand side,

\begin{equation} \label{eq7}
    \int \textrm{log} \left(p_{\theta}(z|\mathcal{T})q_{\phi}(z|\mathcal{S})\right) dz = 2 \textrm{log} \left(1 - D^{2}_{H}\right).
\end{equation}

Noting that $\textrm{log}(p_{\theta}(z|\mathcal{T})q_{\phi}(z|\mathcal{S})) = \textrm{log}(p_{\theta}(z|\mathcal{T})) + \textrm{log}(q_{\phi}(z|\mathcal{S}))$, the left-hand side becomes

\begin{align} \label{eq8}
\begin{split}
    \int \textrm{log}(p_{\theta}(z|\mathcal{T})) dz + \int \textrm{log}(q_{\phi}(z|\mathcal{S})) dz = 2 \textrm{log} \left(1 - D^{2}_{H}\right).
\end{split}
\end{align}

Inserting the positive and negative of the evidence term on the left-hand side:

\begin{align} \label{eq9}
\begin{split}
    \int \textrm{log}(p_{\theta}(z|\mathcal{T})) dz + \int \textrm{log}(q_{\phi}(z|\mathcal{S}))) dz \\ 
    + \int\textrm{log} (p_{\theta}(\mathcal{T}))dz - \int\textrm{log} (p_{\theta}(\mathcal{T}))dz\\
    = 2 \textrm{log} \left(1 - D^{2}_{H}\right),
\end{split}
\end{align}

the following relationship we wanted to arrive at is obtained,

\begin{align} \label{eq10}
\begin{split}
    \int \textrm{log}(p_{\theta}(z,\mathcal{T})) dz + \int \textrm{log}\left(\frac{q_{\phi}(z|\mathcal{S})}{p_{\theta}(\mathcal{T})}\right) dz \\ 
    = 2 \textrm{log} \left(1 - D^{2}_{H}\right),
\end{split}
\end{align}

where $p_{\theta}(z,\mathcal{T}) = p_{\theta}(z|\mathcal{T})p_{\theta}(\mathcal{T})$. Inserting $\frac{q_{\phi}(z|\mathcal{S})}{q_{\phi}(z|\mathcal{S})}$ in the parenthesis of $\textrm{log}(p_{\theta}(z,\mathcal{T}))$ and expanding the logarithm,

\begin{align} \label{eq11}
\begin{split}
    \int \textrm{log}\left(\frac{p_{\theta}(z,\mathcal{T})}{q_{\phi}(z|\mathcal{S})}\right) dz + \int \textrm{log}\left(\frac{q^{2}_{\phi}(z|\mathcal{S})}{p_{\theta}(\mathcal{T})}\right) dz \\ 
    = 2 \textrm{log} \left(1 - D^{2}_{H}\right),
\end{split}
\end{align}

where the second term on the left can be expanded out to obtain

\begin{align} \label{eq12}
\begin{split}
    \int \textrm{log}\left(\frac{p_{\theta}(z,\mathcal{T})}{q_{\phi}(z|\mathcal{S})}\right) dz + 2 \int \textrm{log}(q_{\phi}(z|\mathcal{S})) dz \\ - \int \textrm{log} (p_{\theta}(\mathcal{T})) dz 
    = 2 \textrm{log} \left(1 - D^{2}_{H}\right).
\end{split}
\end{align}

Inserting the fraction $\frac{p_{\theta}(z,\mathcal{T})}{p_{\theta}(z,\mathcal{T})}$ in the parenthesis of $\textrm{log}(q_{\phi}(z|\mathcal{S}))$, the term $2 \int \textrm{log}(q_{\phi}(z|\mathcal{S})) dz$ can be written as

\begin{align} \label{eq13}
\begin{split}
    2 \int \textrm{log}(q_{\phi}(z|\mathcal{S})) dz = 2 \int \textrm{log} ({p_{\theta}(z,\mathcal{T})}) dz \\
    + 2\int \textrm{log} \left(\frac{q_{\phi}(z|\mathcal{S})}{p_{\theta}(z,\mathcal{T})}\right)
\end{split}
\end{align}

in which the second term on the right is the inverse, and consequently, the negative of the twice of the integral of the ELBO logarithmic term, i.e., $-2\int \textrm{log} \left(\frac{p_{\theta}(z,\mathcal{T})}{q_{\phi}(z|\mathcal{S})}\right) dz$. Therefore

\begin{align} \label{eq14}
\begin{split}
    \int \textrm{log}\left(\frac{p_{\theta}(z,\mathcal{T})}{q_{\phi}(z|\mathcal{S})}\right) dz -2\int \textrm{log} \left(\frac{p_{\theta}(z,\mathcal{T})}{q_{\phi}(z|\mathcal{S})}\right) \\
    + 2 \int \textrm{log} ({p_{\theta}(z,\mathcal{T})}) dz - \int \textrm{log} (p_{\theta}(\mathcal{T})) dz \\
    = 2 \textrm{log} \left(1 - D^{2}_{H}\right).
\end{split}
\end{align}

$p_{\theta}(z,\mathcal{T}) = p_{\theta}(z|\mathcal{T})p_{\theta}(\mathcal{T})$ is utilized again, this time on the $\textrm{log} ({p_{\theta}(z,\mathcal{T})})$ term, to obtain

\begin{align} \label{eq15}
\begin{split}
    - \int \textrm{log}\left(\frac{p_{\theta}(z,\mathcal{T})}{q_{\phi}(z|\mathcal{S})}\right) dz  + 2 \int \textrm{log} (p_{\theta}(z|\mathcal{T})p_{\theta}(\mathcal{T})) dz \\ -\int \textrm{log} (p_{\theta}(\mathcal{T})) dz 
    = 2 \textrm{log} \left(1 - D^{2}_{H}\right),
\end{split}
\end{align}

in which the second term can be expanded out to obtain

\begin{align} \label{eq16}
\begin{split}
    - \int \textrm{log}\left(\frac{p_{\theta}(z,\mathcal{T})}{q_{\phi}(z|\mathcal{S})}\right) dz  + 2 \int \textrm{log} (p_{\theta}(z|\mathcal{T})) dz \\ + 2\int \textrm{log} (p_{\theta}(\mathcal{T})) dz -\int \textrm{log} (p_{\theta}(\mathcal{T})) dz 
    = 2 \textrm{log} \left(1 - D^{2}_{H}\right).
\end{split}
\end{align}

Now, using the fact that $p_{\theta}(z|\mathcal{T}) = \frac{p_{\theta}(z,\mathcal{T})}{p_{\theta}(\mathcal{T})}$, the $\textrm{log} ({p_{\theta}(z|\mathcal{T})})$ term can be written as $\textrm{log} \left(\frac{p_{\theta}(z,\mathcal{T})}{p_{\theta}(\mathcal{T})}\right)$, which then along with the other terms, can be expressed as

\begin{align} \label{eq17}
\begin{split}
    - \int \textrm{log}\left(\frac{p_{\theta}(z,\mathcal{T})}{q_{\phi}(z|\mathcal{S})}\right) dz  + 2 \int \textrm{log} \left(\frac{p_{\theta}(z,\mathcal{T})}{p_{\theta}(\mathcal{T})} \right) dz \\ + \int \textrm{log} (p_{\theta}(\mathcal{T})) dz = 2 \textrm{log} \left(1 - D^{2}_{H}\right),
\end{split}
\end{align}

\begin{align} \label{eq18}
\begin{split}
    - \int \textrm{log}\left(\frac{p_{\theta}(z,\mathcal{T})}{q_{\phi}(z|\mathcal{S})}\right) dz  + 2 \int \textrm{log} \left(p_{\theta}(z,\mathcal{T}) \right) dz \\ - \int \textrm{log} (p_{\theta}(\mathcal{T})) dz = 2 \textrm{log} \left(1 - D^{2}_{H}\right),
\end{split}
\end{align}

\begin{align} \label{eq19}
\begin{split}
     - \int \textrm{log}\left(\frac{p_{\theta}(z,\mathcal{T})}{q_{\phi}(z|\mathcal{S})}\right) dz  + 2 \int \textrm{log} \left(p_{\theta}(z,\mathcal{T}) \right) dz   = \\\int \textrm{log} (p_{\theta}(\mathcal{T})) dz  + 2 \textrm{log} \left(1 - D^{2}_{H}\right).
\end{split}
\end{align}

Finally, noting the observation that $\int \textrm{log}\left(\frac{p_{\theta}(z,\mathcal{T})}{q_{\phi}(z|\mathcal{S})}\right) dz$ can be rewritten in terms of ELBO and $q_{\phi}(z|\mathcal{S})$ as

\begin{align} \label{eq20}
\begin{split}
    \int \textrm{log}\left(\frac{p_{\theta}(z,\mathcal{T})}{q_{\phi}(z|\mathcal{S})}\right) dz = \int \left(\frac{1}{q_{\phi}(z|\mathcal{S})}\frac{d(ELBO)}{dz} \right) dz \\ = \frac{1}{q_{\phi}(z|\mathcal{S})}ELBO, 
\end{split}
\end{align}

the following is derived:

\begin{align} \label{eq21}
\begin{split}
   ELBO' =  \int \textrm{log} (p_{\theta}(\mathcal{T})) dz +  \textrm{log} \left(1 - D^{2}_{H}\right)^{2}
\end{split}
\end{align}

where $ELBO' = \frac{ELBO}{q_{\phi}(z|\mathcal{S})}$, $\int \textrm{log} (p_{\theta}(\mathcal{T}))dz$ is the \emph{evidence} term, with the new ELBO term, $ELBO'$ written as $\frac{ELBO}{q_{\phi}(z|\mathcal{S})}$. For comparison purposes, we included the ELBO for variational model utilizing the KL divergence:

\begin{equation} \label{eq22}
\begin{split}
    ELBO = \int \textrm{log} (p_{\theta}(\mathcal{T}))dz - D_{KL}(q_{\phi}(z|\mathcal{S})||p_{\theta}(z|\mathcal{T}))),\\ 
\end{split}
\end{equation}

and we can see that for $ELBO'$, not only the square of the logarithm of the difference in $D^{2}_{H}$ with respect to 1 is needed, which ensured that the range of $D^{2}_{H}$, and thus $D_{H}$ lies in $[0,1]$, the $ELBO'$ appeared as the linear sum of the evidence and the logarithm of the square of the parenthesis containing the $D_{H}$, rather than the linear differences between the evidence and the relevant divergence (as in the case of $D_{KL}$). Nevertheless, since $\textrm{log}\left(1 - D^{2}_{H}\right)^{2}$ have the range $(-\infty,0]$, while $D_{KL}(q_{\phi}(z|\mathcal{S})||p_{\theta}(z|\mathcal{T})))$ have the range $[0,\infty)$, if both terms approaches zero (which occurs when $q_{\phi}(z|\mathcal{S}) = p_{\theta}(z|\mathcal{T})$), the respective ELBO values equate to the evidence. Therefore, maximizing the evidence term is equivalent to maximizing the respective lower bounds.


We may describe $D_{H}$ in terms of the means and standard deviations of the prototypes of the $i$\textsuperscript{th} class in the prior distribution ($\mu_{i},\Sigma_{i}$) and $j$\textsuperscript{th} class in the posterior distribution ($\mu_{j},\Sigma_{j}$), similar to that of the KL divergence. We merely state the mathematical form here, which is a Hellinger distance between two Gaussian distributions and is obtained in \cite{pardo2018statistical}:


\begin{equation} \label{eq23}
\begin{split}
    D^{2}_{H} = 1 - \frac{\det{(\Sigma_{i})}^{1/4}\det{(\Sigma_{j})}^{1/4}}{\det{\left(\frac{\Sigma_{i}+\Sigma_{j}}{2}\right)}^{1/2}} \textrm{exp}\left(-\frac{1}{8}(\mu_{i} - \mu_{j})^{T}\left(\frac{\Sigma_{i}+\Sigma_{j}}{2}\right)^{-1}(\mu_{i} - \mu_{j})\right).\\ 
\end{split}
\end{equation}

If one of the distribution can be described by a normal distribution, i.e., $\mu_{j} = 0$, $\Sigma_{j} = 1$, then $D^{2}_{H}$ can be reduced to

\begin{equation} 
\begin{split}
    D^{2}_{H} = 1 - \frac{\det{(\Sigma_{i})}^{1/4}}{\det{\left(\frac{\Sigma_{i}+1}{2}\right)}^{1/2}} \textrm{exp}\left(-\frac{1}{8}(\mu_{i})^{T}\left(\frac{\Sigma_{i}+1}{2}\right)^{-1}(\mu_{i})\right).\\ 
\end{split}
\end{equation}

This mathematical form is utilized in our $ELBO'$ optimization for image reconstruction. Furthermore, the Hellinger distance can also be thought of as a generalization of the Mahalanobis distance $D_{M}$ \cite{mclachlan1999mahalanobis}, and this is because $D_{H}$ is related to the Bhattarcharyya distance $D_{B}$ \cite{bhattacharyya1946measure}, which is another $f$-divergence metric, via the Bhattarcharyya coefficient $BC$ as $BC = \sqrt{1-D_{H}^{2}}$. This distance, in turn, can be associated with $D_{M}$ \cite{kashyap2019perfect} via

\begin{equation}
    D_{B} = -\textrm{ln}(BC) = \frac{1}{8}(D_{M}) + \frac{1}{2}\textrm{ln}\left(\frac{|\Sigma_{i} + \Sigma_{j}|/2}{|\Sigma_{j}|^{1/2}|\Sigma_{2}|^{1/2}}\right).
\end{equation}

Incorporating $D_{H}$ allows our model to attain new level of few-shot classification performances as it is able to account for outlier samples which could be encountered during the few-shot sampling. Such outliers could also be common due to the high intra-class similarity presented in images of certain few-shot datasets during the feature clustering.

Although Equation~(21) reveals how to rewrite the ELBO in terms of the Hellinger distance, this change also affects the optimization landscape and the stability of gradient-based methods. Unlike unbounded divergences (e.g., KL), the Hellinger distance 
$D_H$ lies in $[0,1]$, and therefore $\log(1 - D_H^2)$ in \eqref{eq21} remains in $(-\infty, 0]$. This boundedness curbs the risk of large or “explosive” gradients when $p_\theta(z \mid T)$ and $q_\phi(z \mid S)$ partially overlap. Concretely: 
\begin{itemize}
    \item \textbf{Convergence Behavior.} Because $\log(1 - D_H^2)$ has a smoother gradient profile near $D_H^2 = 1$ compared to unbounded divergences, gradient steps tend to be more uniform across episodes with varying degrees of overlap. In few-shot scenarios, where each meta-batch can have large distribution mismatches, sharper or unbounded gradients can cause instability. Hence, Hellinger-based updates often converge more reliably.
    \item \textbf{Mitigation of Gradient Spikes.} When the training distribution $q_\phi(z \mid S)$ is far from the prior $p_\theta(z \mid T)$, unbounded divergences (e.g., KL) may produce steep gradient norms. By contrast, $D_H$'s ceiling (unity) forces $\log(1 - D_H^2)$ to saturate in a narrower range, thereby limiting runaway gradient magnitudes and ensuring more stable SGD steps.
    \item \textbf{Symmetry Benefits.} Hellinger distance is symmetric, i.e., $D_H(p, q) = D_H(q, p)$. During meta-training, this property guarantees that the “direction” of distribution comparisons remains consistent even if we swap $p_\theta(z \mid T)$ and $q_\phi(z \mid S)$. Empirically, we have observed more consistent parameter updates across tasks that vary in how the overlap occurs.
\end{itemize}

In practice, these properties manifest as smoother training curves and fewer sudden drops or spikes in accuracy. They also lead to faster or more consistent convergence in stochastic gradient-based schemes, particularly under the data-scarce conditions that define few-shot learning.

\subsubsection{Theoretical Foundations of Hellinger Distance Selection}

The selection of Hellinger distance as our core metric is motivated by several key theoretical advantages over other symmetrical distances like the Wasserstein distance. While both distances offer symmetry properties, the Hellinger distance $D_H^2(P,Q) = 1 - \int\sqrt{p(x)q(x)}dx$ provides unique benefits particularly suited to few-shot learning scenarios. First, it offers closed-form solutions for many probability distributions, including Gaussian distributions where $D_H^2 = 1 - \frac{\det(\Sigma_1)^{1/4}\det(\Sigma_2)^{1/4}}{\det\left(\frac{\Sigma_1+\Sigma_2}{2}\right)^{1/2}}\exp\left(-\frac{1}{8}(\mu_1-\mu_2)^T\left(\frac{\Sigma_1+\Sigma_2}{2}\right)^{-1}(\mu_1-\mu_2)\right)$. This contrasts with the Wasserstein distance $W_p(P,Q) = \left(\inf_{\gamma \in \Gamma(P,Q)} \int\int d(x,y)^p d\gamma(x,y)\right)^{1/p}$, which requires solving an optimization problem. Second, as an f-divergence measure, the Hellinger distance is naturally bounded between [0,1], providing better numerical stability and more interpretable similarity measures compared to unbounded metrics. This property is particularly valuable in few-shot scenarios where limited samples can lead to unstable distance estimates and where outlier samples can significantly disrupt training. Third, the Hellinger distance can be interpreted geometrically as the $L_2$ norm of square-rooted differences between probability densities: $D_H(P,Q) = \frac{1}{\sqrt{2}}\|\sqrt{p}-\sqrt{q}\|_2$, facilitating more stable optimization in prototype-based few-shot methods. Fourth, the Hellinger distance's relationship with the KL-divergence ($D_H^2(P,Q) \approx \frac{1}{4}D_{KL}(P||Q)$ for small differences) provides theoretical guarantees while maintaining symmetry. This relationship enables us to rewrite the ELBO objective traditionally expressed with $D_{KL}$ into an expression involving $\log(1 - D_H^2)$, which remains in $(-\infty, 0]$, enforcing numerical stability. Finally, its gradient form $\nabla_\theta D_H^2(p_\theta||q_\phi) = -\frac{1}{2}\int\sqrt{\frac{q_\phi}{p_\theta}}\nabla_\theta p_\theta dx$ leads to more stable optimization in the few-shot setting where sample efficiency is crucial. This stability is particularly important when dealing with sparse distributions in query classes, as the bounded nature prevents gradient explosion while maintaining reliable updates during meta-training. Our empirical results (Table 7) further support this choice, showing that our Hellinger-based approach achieves superior reconstruction quality (FID = 2.75) compared to Wasserstein-based methods (FID = 3.38). This combination of theoretical properties - efficient computation, natural bounds, geometric interpretability, robust gradients, and direct optimization formulation - along with empirical validation makes the Hellinger distance particularly well-suited for our few-shot learning framework.

\subsection{ANROT-HELANet Training}

For the encoder feature extraction, we utilized the ResNet12 backbone \cite{he2016deep}. In our training approach, the support $\mathcal{S}$ and target query set $\mathcal{Q}$ are partitioned into $N$ subset, each belonging to a certain class (i.e., $\mathcal{S}_{1}$, $\mathcal{S}_{2}$, ..., $\mathcal{S}_{N}$). Hence there exists $N$ posterior class-specific distribution, all conditioned on $\mathcal{S}$, which are utilized for the estimation. The distributions all satisfies a Gaussian with a diagonal covariance structure $\mathcal{N}(\mu, \sigma^{2})$, where the mean $\mu$
and standard deviation $\sigma$ are a set of class distributed values. To address the issue of intractable backpropagation due to the stochastic sampling required to arrive at the latent variable $z$,  the reparameterization trick \cite{kingma2015variational} is employed ($z = \mu + \sigma \odot \zeta$), where $\zeta$ are random variables generated from a standard normal distribution $\mathcal{N}(0,1)$, and $\odot$ represents the Hadamard product. 

The probability of the target data samples that belongs to their respective classes are then computed and the maximum values obtained serves as the predicted label. This is then compared to the ground-truth label $y_{i}$ using the categorical cross entropy loss $\mathcal{L}_{CCE}$:

\begin{equation} \label{eq24}
    \mathcal{L}_{CCE} = -\sum_{i=1}^{Q} y_{i} \textrm{log} (p(\hat{y_{i}} = y_{i}| \mathcal{T}_{meta}))
\end{equation}

where $Q$ denotes the number of query samples in a batch-trained meta task, denoted by $\mathcal{T}_{meta}$. 

Our proposed Hellinger Similarity softmax loss function $\mathcal{L}_{Hesim}$ is motivated by the cosine similarity softmax loss in the contrastive similarity loss functions. The original cosine similarity can be thought of as the dot product between the class-relevant feature vector and the class prototype. Our $\mathcal{L}_{Hesim}$  modification lies in replacing the similarity function in the latter with the Hellinger similarity. Since the Hellinger distance is the probabilistic analog of the feature vectors computation in a Euclidean distance-like manner, we can compute the class-specific distributions of the query vectors $v_{(Q,i)}$ along with the prototype vectors $v_{(P,i)}$, which is followed by computing its probability of belonging to the k\textsuperscript{th}-class. These mathematical operation is culminated in our loss function as

\begin{equation} \label{eq25}
    \mathcal{L}_{Hesim} = - \sum_{i=1}^{N_{t}}y_{i}\textrm{log} (p(y=j|v_{(Q,i)}))
\end{equation}

in which $N_{t}$ denotes the number of training samples. $p(y=j| v_{(Q,i)})$ is then computed from 

\begin{equation} \label{eq26}
    p(y=j|v_{(Q,i)}) = \frac{\textrm{exp}(-\{v_{(Q,i)}, c_{(Q,j)}\})}{\sum_{i=1}^{N} \textrm{exp} \left(-\{v_{(Q,i)}, c_{(Q,j)}\}\right)}
\end{equation}

where $c_{(Q,j)}$ represents the j\textsuperscript{th}-class prototype from query set. Observe that in the denominator, the sum over the total number of classes $N$ is taken. The Hellinger similarity is contained in the term $\{v_{(Q,i)}, c_{(Q,j)}\}$ in the parenthesis of $p(y=j| v_{(Q,i)})$ and consequently, $\mathcal{L}_{Hesim}$. 

Similar to prior related works, the other loss required includes the image reconstruction loss $\mathcal{L}_{rec}$ which guides the image reconstruction process in the decoder component $\mathcal{F}_{dec}$ of our network. It is mathematically described as the L1-norm between the image input $I$ and the reconstructed support image $I'$.

\begin{equation} \label{eqn27}
    \mathcal{L}_{rec} = ||\mathcal{F}_{dec}(z) - I || = ||I' - I||.
\end{equation}

\begin{figure*}[hbt!]
    \centering
    \includegraphics[scale=0.53]{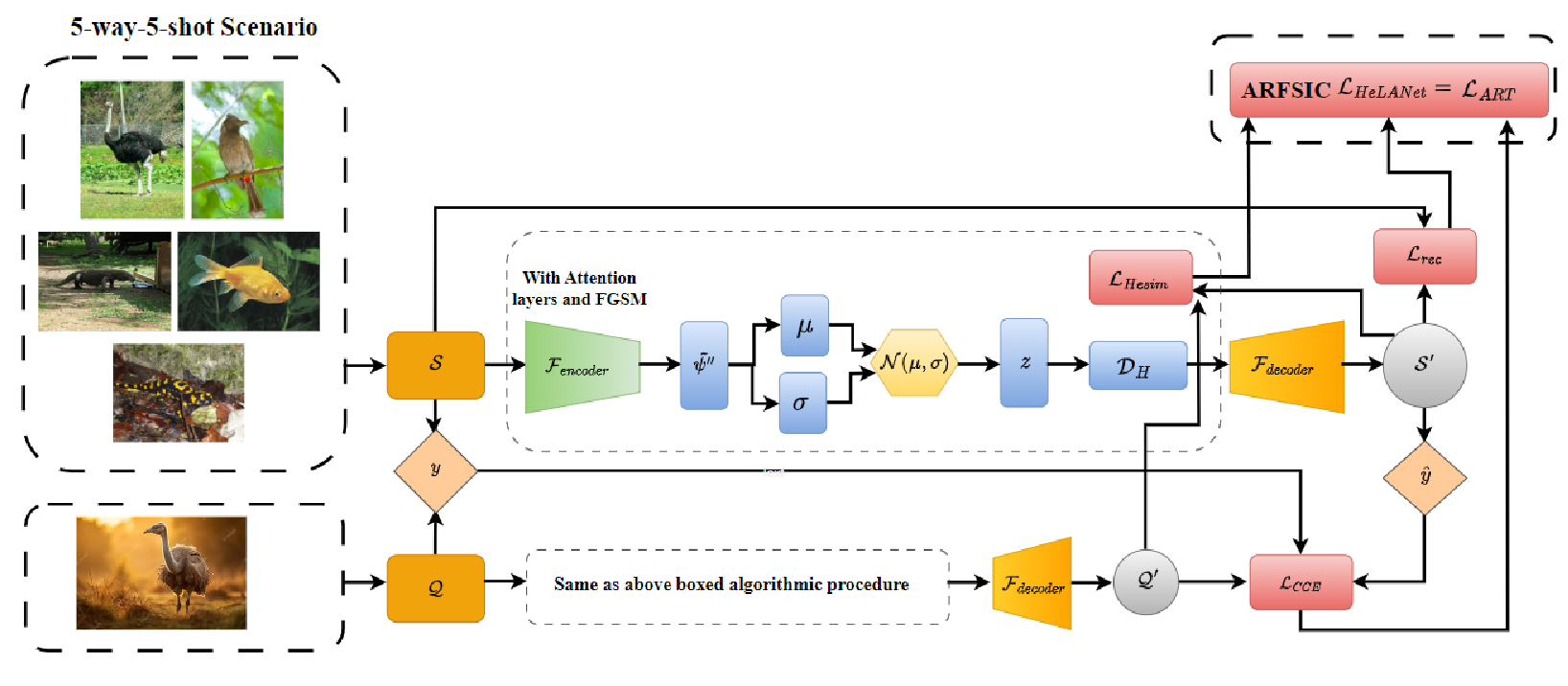}
    \caption{Algorithmic architecture of our overall ANROT-HELANet for a 5-way-5-shot scenario.}
    \label{ART_HELANET2}
\end{figure*}

In summary, our HELANet (without the adversarial and natural robust procedure) training loss, $\mathcal{L}_{HELANet}$, is

\begin{equation} \label{eq28}
    \mathcal{L}_{HELANet} = \mathcal{L}_{CCE} + \lambda_{1} \cdot \mathcal{L}_{Hesim} + \lambda_{2} \cdot \mathcal{L}_{rec},
\end{equation}

where $\lambda_{1}$, $\lambda_{2}$ denotes the weight parameters of $\mathcal{L}_{Hesim}$ and $\mathcal{L}_{rec}$, with respect to $\mathcal{L}_{CCE}$. The value of $\lambda_{1}$ is set as 0.5 while the value of $\lambda_{2}$ is set as 1.0. To achieve adversarial robustness, $\mathcal{L}_{HELANet}$ is subjected to a minimax optimization \cite{madry2017towards}, which is mathematically depicted as

\begin{equation} \label{eq29}
    \mathcal{L}_{ANROT} = \min_{\theta} \EX_{(\bf{x},y) \sim \mathcal{D}}\left[\max_{||\delta||_{p} < \epsilon} \mathcal{L}_{HELANet} (\bf{x} + \delta, y)_{\theta}\right]
\end{equation}

where $\delta$ is the adversarial perturbation. Therefore our ANROT-HELANet loss function, $\mathcal{L}_{ANROT}$, consisted of searching for the optimal parameter $\theta$ in the configuration that retains the lowest loss, despite the addition of the perturbations. The algorithmic architecture of our ANROT-HELANet is illustrated in Figure \ref{ART_HELANET2} for a 5-way-5-shot scenario, while its algorithm outline is presented in Algorithm 1.

\begin{algorithm}
\caption{Algorithmic outline of ANROT-HELANet}\label{algo1}
\textbf{Input}: Original support image set $\mathcal{S}$ and labels $y_{\mathcal{S}}$ in the meta-training set, query image set $\mathcal{Q}$ and labels $y_\mathcal{Q}$ in the meta-training set, corrupted support and query image set $\mathcal{S'}_{g}$, $\mathcal{Q'}_{g}$ by Gaussian noise in the meta-training set, number of classes $N_{c}$. Encoder $f$ and decoder $d$. \\
\textbf{Output}: Predicted query labels $\hat{y}_\mathcal{Q}$ in the meta-test set, restored query and support image samples $\mathcal{\tilde{Q}}$, $\mathcal{\tilde{S}}$ in the meta-test dataset. 
\begin{algorithmic}[1]
\FOR{batch in meta-train data}
\STATE Choose a subset class $C$ from $N_{c}$ classes in the meta-training dataset randomly.
\STATE  Initialize loss functions $\mathcal{L}_{CCE}$, $\mathcal{L}_{rec}$, and $\mathcal{L}_{Hesim}$ as zero.
\FOR {$C$ in $N_{c}$}
\STATE  Select number $N_{\mathcal{S}}$ and $N_{\mathcal{Q}}$ of support and query examples respectively for $C$ from batch. Also select number $N_{\mathcal{S'}}$, $N_{\mathcal{Q'}}$ of corrupted support and query examples due to the gaussian noise from the same batch respectively. 
\STATE  Perform embedding of $\mathcal{S}_{C}$ and $\mathcal{Q}_{C}$ as $f(\mathcal{S}_{C})$ and $f(\mathcal{Q}_{C})$. The corresponding gaussian corrupted embedded features are embedded as $f(\mathcal{S'}_{C})$ and $f(\mathcal{Q'}_{C})$. At the same time the corresponding adversarial corrupted embedded features are created via equation \ref{eq3}.
\STATE  Compute class prototypes as $c = \frac{1}{N_{c}}\sum_{\subset{\mathcal{S_{C}}}}f(\mathcal{S}_{C})$. Do the corresponding for the adversarial and gaussian corrupted embedded features. 
\STATE  Use reparameterization trick $z = \mu + \sigma \odot \zeta$ to obtain the query and support latent features $z_{\mathcal{S}_{C}}$ and $z_{\mathcal{S}_{Q}}$. Again do the same for the adversarial and gaussian corrupted embedded features. 
\ENDFOR
\FOR{$\mathcal{Q}$ in batch query points}
\FOR{$C$ in $N_{c}$}
\STATE Compute Hellinger distances $D_{H}$ between the respective embedded query features and respective class prototypes distribution as given in equation \ref{eq4}.
\STATE Update the Hellinger Similarity loss function $\mathcal{L}_{Hesim}$.
\STATE Use decoder $d(z_{\mathcal{Q}_{c}})$ to reconstruct query image $\tilde{\mathcal{Q}}$, update $\mathcal{L}_{rec}$.
\ENDFOR
\STATE The predicted label is cast as the minimum value of $D_{H}$ between prototypes and embedded query features $\hat{y_{Q}} \rightarrow \underset{x}{\mathrm{argmin}} D_{H}$.
\STATE Update the cross entropy loss function $\mathcal{L}_{CCE}$.
\ENDFOR
\ENDFOR
\STATE \textbf{return} $\hat{y}_{Q}$, $\mathcal{\tilde{Q}}$, $\mathcal{\tilde{S}}$, either which described the $I'$ in equation \ref{eqn27}. 
\end{algorithmic}
\end{algorithm}

\section{Experiments}

We utilized the benchmarked FC-100, CIFAR-FS, miniImageNet, and tieredImageNet dataset for evaluating our approach relative to the SOTAs FSL techniques, which are the same as that listed out in \cite{roy2022felmi}. The code repository for our work is available at https://github.com/GreedYLearner1146/ANROT-HELANet/tree/main.

\subsection{FC-100 and CIFAR-FS}
As a subset of the CIFAR-100 dataset, the FC-100 contains 60 meta-training classes, 20 meta-validation classes, and 20 meta-testing classes. There are 600 pictures in overall, each of image size 32 $\times$ 32. Few-shot learning can be rendered a challenge in this dataset due to the low image quality and small image size. The CIFAR-FS dataset, which contains 64 meta-training classes, 16 meta-validation classes, and 20 meta-testing classes, is also a subset of the CIFAR-100 dataset. Like FC-100, the image size is of 32 $\times$ 32, and once again few-shot learning execution can be challenging due to the significant intra-class similarity between some of the images. 

\subsection{miniImageNet and tieredImageNet}
The miniImageNet dataset is a subset of the ImageNet dataset \cite{deng2009imagenet} includes a broad range of images extracted from 100 classes, with 600 images per class, each of size 84 $\times$ 84. There are 64 meta-training classes, 16 meta-validation classes, and 20 meta-testing classes. The tieredImageNet is also a subset of the ImageNet, but utilized more classes for the train-valid-test split, more specifically 351 classes for meta-training, 97 classes for meta-validation, and 160 classes for meta-testing.

\subsection{Settings}
All experiments were conducted using the Tesla A100 Graphical Processing Units (GPU) from Google Colab, with PyTorch and Tensorflow serving as the underlying libraries. We reported the few-shot classification accuracy (in $\%$) for our technique, using the $N$-way-1-shot and $N$-way-5-shot evaluation approach, in line with a majority of FSL literatures. We have also leveraged the respective data augmentation that is implemented in many few-shot literatures catered to the respective dataset images.

\section{Experimental Results}

\subsection{Main Findings}

\begin{figure*}[hbt!]
    \centering
    \includegraphics[scale = 0.80]{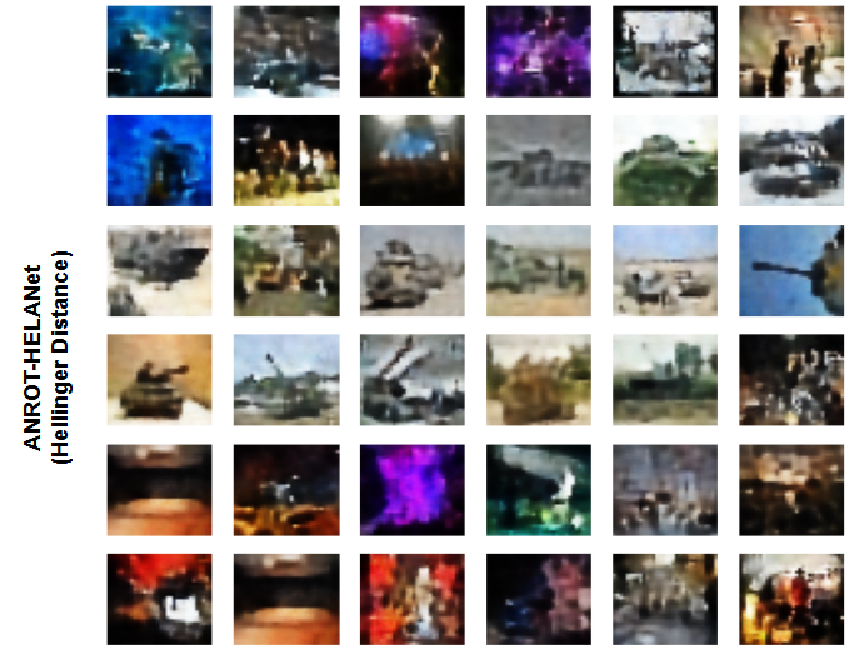}
    \caption{The miniImageNet reconstructed images obtained from the Vanilla  VAE architectures using the Hellinger distance. (\textbf{zoom in for the best view}).}
    \label{fig:Fig4}
\end{figure*}

\begin{table}
\caption{The classification accuracy (in $\%$) using the 5-way 1-shot and 5-way 5-shot learning evaluation for our ANROT-HELANet relative to the SOTAs on the \textbf{CIFAR-FS}. All CNN backbone used is the ResNet-12.}
\begin{tabular}{p{5cm}p{3cm}p{3cm}}
\hline
\textbf{Methods} & \textbf{5-way-1-shot} & \textbf{5-way-5-shot} \\
\hline
 ProtoNet \cite{snell2017prototypical} & 72.7$\pm$0.7 &  83.5$\pm$0.5 \\
 MetaOptNet \cite{lee2019meta} & 72.6$\pm$0.7 & 84.3$\pm$0.5  \\
 DSN-MR \cite{simon2020adaptive} & 75.6$\pm$0.9 & 86.2$\pm$0.6 \\
 RFS-Simple \cite{tian2020rethinking} & 71.5$\pm$0.8 & 86.0$\pm$0.5 \\
 RFS-distill \cite{tian2020rethinking}  & 73.9$\pm$0.8 & 86.9$\pm$0.5 \\
 IER-distill \cite{rizve2021exploring}  & 77.6$\pm$1.0 & 89.7$\pm$0.6  \\
 PAL \cite{ma2021partner} & 77.1$\pm$0.7 & 88.0$\pm$0.5 \\
 SKD-Gen1 \cite{rajasegaran2020self} & 76.6$\pm$0.9 & 88.6$\pm$0.5 \\
 Label Halluc \cite{jian2022label} & 78.0$\pm$1.0 & 89.4$\pm$0.6\\
 FeLMi \cite{roy2022felmi} & 78.2$\pm$0.7 & 89.5$\pm$0.5 \\
 TRIDENT \cite{singh2022transductive} & 78.1$\pm$1.2 & 88.3$\pm$0.8 \\
 HELA-VFA \cite{lee2024hela} & 78.9$\pm$0.4 & 90.7$\pm$0.7 \\
 \textbf{ANROT-HELANet (Ours)} & \bf{79.2$\pm$0.7} & \bf{90.9$\pm$0.5} \\
 \hline
\end{tabular}
\label{tab:CIFAR-FS FSL}
\end{table}

\begin{table}
\caption{The classification accuracy (in $\%$) using the 5-way 1-shot and 5-way 5-shot learning evaluation for our ANROT-HELANet relative to the SOTAs on the \textbf{FC-100}. All CNN backbone used is the ResNet-12, except for AssoAlign which utilized a ResNet-18 backbone.}  
\begin{tabular}{p{5cm}p{3cm}p{3cm}}
\hline
\textbf{Methods} & \textbf{5-way-1-shot} & \textbf{5-way-5-shot} \\
\hline
 ProtoNet \cite{snell2017prototypical}  & 37.5$\pm$0.6 &  52.5$\pm$0.6 \\
 MetaOptNet \cite{lee2019meta} & 41.1$\pm$0.6 & 55.5$\pm$0.6  \\
 TADAM \cite{oreshkin2018tadam} & 40.1$\pm$0.4 & 56.1$\pm$0.4 \\ 
 MTL \cite{sun2019meta} & 45.1$\pm$1.8 & 57.6$\pm$0.9 \\
 RFS-Simple \cite{tian2020rethinking} & 42.6$\pm$0.7 & 59.1$\pm$0.6 \\
 Deep-EMD \cite{zhang2020deepemd} & 46.5$\pm$0.8 & 63.2$\pm$0.7 \\
 RFS-distill \cite{tian2020rethinking} & 44.6$\pm$0.7 & 60.9$\pm$0.6 \\
 IER-distill \cite{rizve2021exploring} & 48.1$\pm$0.8 & 65.0$\pm$0.7  \\
 PAL \cite{ma2021partner} & 47.2$\pm$0.6 & 64.0$\pm$0.6 \\
 SKD-Gen1 \cite{rajasegaran2020self}  & 46.5 $\pm$0.8 & 64.2$\pm$0.8 \\
 AssoAlign \cite{afrasiyabi2020associative} & 45.8 $\pm$0.5 & 59.7$\pm$0.6 \\
 InfoPatch \cite{gao2021contrastive} & 43.8 $\pm$0.4 & 58.0$\pm$0.4 \\
 Label Halluc \cite{jian2022label} & 47.3$\pm$0.7 & 67.9$\pm$0.7\\
 FeLMi \cite{roy2022felmi} & 49.0$\pm$0.7 & 68.7$\pm$0.7 \\
 TRIDENT \cite{singh2022transductive} & 48.4$\pm$0.7 & 66.2$\pm$0.4\\
 HELA-VFA \cite{lee2024hela} & 50.3$\pm$0.3 & 69.1$\pm$0.2  \\
 \textbf{ANROT-HELANet (Ours)} & \bf{51.2$\pm$0.6} & \bf{69.6$\pm$0.6}  \\
 \hline
\end{tabular}
\label{tab:FC-100 FSL}
\end{table}

\begin{table} 
\caption{The classification accuracy (in $\%$) evaluations on the \textbf{miniImageNet}. All CNN backbone used is the ResNet-12, except for AssoAlign which utilized a ResNet-18 backbone.}
\begin{tabular}{p{5cm}p{3cm}p{3cm}}
\hline
\textbf{Methods} & \textbf{5-way-1-shot} & \textbf{5-way-5-shot} \\
\hline
 ProtoNet \cite{snell2017prototypical} & 60.4$\pm$0.8 & 78.0$\pm$0.6\\
 MetaOptNet \cite{lee2019meta} & 62.6$\pm$0.6 & 78.6$\pm$0.5 \\
 MTL \cite{sun2019meta} & 61.2$\pm$1.8 & 75.5$\pm$0.8\\
 TADAM \cite{oreshkin2018tadam} & 58.5$\pm$0.3 & 76.7$\pm$0.3\\
 Shot-Free \cite{ravichandran2019few} & 59.0$\pm$0.4 & 77.6$\pm$0.4\\
 Deep-EMD \cite{zhang2020deepemd} & 65.9$\pm$0.8 & 82.4$\pm$0.6\\
 FEAT \cite{ye2020few} & 66.8$\pm$0.2 & 82.1$\pm$0.1\\
 DSN-MR \cite{simon2020adaptive} & 64.6$\pm$0.7 & 79.5$\pm$0.5\\
 Neg-Cosine \cite{liu2020negative}  & 63.9$\pm$0.8 & 81.6$\pm$0.6\\
 P-Transfer \cite{shen2021partial}  & 64.2$\pm$0.8 & 80.4$\pm$ 0.6\\
 MELR \cite{fei2020melr}  & 67.4$\pm$0.4 & 83.4$\pm$0.3\\
 TapNet \cite{yoon2019tapnet}  & 61.7$\pm$0.2 & 76.4$\pm$0.1\\
 IEPT \cite{zhang2020iept}  & 67.1$\pm$0.4 & 82.9$\pm$0.3\\
 RFS-Simple \cite{tian2020rethinking}  & 62.0$\pm$0.6 & 79.6$\pm$0.4\\
 RFS-distill \cite{tian2020rethinking} & 64.8$\pm$0.8 & 82.4$\pm$0.4\\
 IER-distill \cite{rizve2021exploring} & 66.9$\pm$0.8 & 84.5$\pm$0.5\\
 SKD-Gen1 \cite{rajasegaran2020self} & 66.5$\pm$1.0 & 83.2$\pm$0.5\\
 AssoAlign \cite{afrasiyabi2020associative} & 60.0$\pm$0.7 & 80.4$\pm$0.7\\
 Label Halluc \cite{jian2022label} & 67.0$\pm$0.7 & 85.9$\pm$0.5\\
 FeLMi \cite{roy2022felmi} & 67.5$\pm$0.8 & 86.1$\pm$0.4\\
 TRIDENT \cite{singh2022transductive} & \bf{86.1$\pm$0.6} & \bf{96.0$\pm$0.3} \\
 HELA-VFA \cite{lee2024hela} & 68.2$\pm$0.3 & 86.7$\pm$0.7\\
 ANROT-HELANet (Ours) & 69.4$\pm$0.3 & 88.1$\pm$0.4\\
 \hline
\end{tabular}
\label{tab:miniImageNetFSL}
\end{table}

\begin{table} 
\caption{The classification accuracy (in $\%$) evaluations on the \textbf{tieredImageNet}. All CNN backbone used is the ResNet-12, except for AssoAlign which utilized a ResNet-18 backbone.}
\begin{tabular}{p{5cm}p{3cm}p{3cm}}
\hline
\textbf{Methods} & \textbf{5-way-1-shot} & \textbf{5-way-5-shot} \\
\hline
 ProtoNet \cite{snell2017prototypical} & 65.7$\pm$0.9 & 83.4$\pm$0.7 \\
 MetaOptNet \cite{lee2019meta} & 66.0$\pm$0.7 & 81.6$\pm$0.5 \\
 MTL \cite{sun2019meta} & 65.6$\pm$1.8 & 80.6$\pm$0.9 \\
 Shot-Free \cite{ravichandran2019few} & 66.9$\pm$0.4 & 82.6$\pm$0.4\\
 Deep-EMD \cite{zhang2020deepemd} & 71.2$\pm$0.9 & 86.0$\pm$0.6 \\
 FEAT \cite{ye2020few} & 70.8$\pm$0.2 & 84.8$\pm$0.2\\
 DSN-MR \cite{simon2020adaptive} & 67.4$\pm$0.8 & 82.9$\pm$0.6\\
 MELR \cite{fei2020melr} & 72.1$\pm$0.5 & 87.0$\pm$0.4 \\
 TapNet \cite{yoon2019tapnet} & 63.1$\pm$0.2 & 80.3$\pm$0.1\\
 IEPT \cite{zhang2020iept}  & 72.2$\pm$0.5 & 86.7$\pm$0.3 \\
 RFS-Simple \cite{tian2020rethinking} & 69.7$\pm$0.7 & 84.4$\pm$0.6\\
 RFS-distill \cite{tian2020rethinking} & 71.5$\pm$0.7 & 86.0$\pm$0.5 \\
 IER-distill \cite{rizve2021exploring} & 72.7$\pm$0.9 & 86.6$\pm$0.8\\
 SKD-Gen1 \cite{rajasegaran2020self} & 72.4 $\pm$1.2 & 86.0$\pm$0.6\\
 AssoAlign \cite{afrasiyabi2020associative} & 69.3$\pm$0.6 & 86.0$\pm$0.5\\
 Label Halluc \cite{jian2022label} & 72.0$\pm$0.9 & 86.8$\pm$0.6\\
 FeLMi \cite{roy2022felmi} & 71.6$\pm$0.9 & 87.1$\pm$0.6\\
 TRIDENT \cite{singh2022transductive} & \bf{87.0$\pm$0.6} & \bf{97.0$\pm$0.2}\\
 HELA-VFA \cite{lee2024hela} & 72.5$\pm$0.5 & 87.6$\pm$0.1 \\
 ANROT-HELANet (Ours) & 75.3$\pm$0.2 & 89.5$\pm$0.8\\
 \hline
\end{tabular}
\label{tab:tieredImageNetFSL}
\end{table}

We compared the performance of our method relative to the selected FSL techniques in each dataset in the same manner as laid out in \cite{roy2022felmi}. We also included two more recent methods (TRIDENT and HELA-VFA \cite{lee2024hela}, both which are variational-based few-shot approaches like our proposed model). Table 2 to 5 illustrate the results of the 5-way-1-shot and 5-way-5-shot evaluations for our approach relative to the selected methods on each of the dataset. All of the chosen techniques employed ResNet-12 as their encoder backbone except for the AssoAlign. The number of training episode is chosen to be 80000 for our model, and the number of validation tasks is set to 100. The learning rate throughout the training is 0.001. Also, for all the results reported, the adversarial and natural robustness training are done utilizing both $\epsilon = 0.05$ and $\sigma = 0.05$ (since the case of having no adversarial or natural gaussian noise just corresponds to our previous HELA-VFA model). One obvious but interesting finding is that for all approaches, the accuracy reported from the 5-way-5-shot evaluations always outperforms the reported values from the 5-way-1-shot evaluations. As there are fewer data instances in the support set to help the model make accurate predictions during each training batch in the latter, employing few-shot learning becomes more challenging in this context. Another interesting finding is that, when compared to CIFAR-FS, the accuracies obtained by both 5-way-1-shot and 5-way-5-shot assessments are lower for FC-100. This suggests that, relative to CIFAR-FS, the FC-100 poses a greater difficulty for FSL to provide accurate classifications. The accuracies computed for the miniImageNet and tieredImageNet suggested that both posed the most difficulties with few-shot categorization. Among these, tieredImageNet is a slightly more challenging than miniImageNet. Almost all datasets show that our ANROT-HELANet method regularly outperforms other SOTA techniques (except for miniImageNet and tieredImageNet where TRIDENT reported more superior numerical metrics). The reason for TRIDENT's high few-shot performances in this two context is due to transductive feature extractor being utilized, meaning that the TRIDENT network decoupled label and semantic information simultaneously using two variational network instead of one. For our ANROT-HELANet, as promised, improvements in the classification accuracy has been observed for all datasets relative to our previous HELA-VFA works by 0.30$\%$ and 0.20$\%$ in the 1-shot and 5-shot scenario for CIFAR-FS, by 0.90$\%$ and 0.50$\%$ for the 1-shot and 5-shot scenario for FC-100, by 1.20$\%$ and 1.40$\%$ for the 1-shot and 5-shot scenario for miniImageNet, and lastly by 2.80$\%$ and 1.90$\%$ for the 1-shot and 5-shot scenario for tieredImageNet. Our simulation also emphasized the importance of data augmentation in the pre-processing steps, as these generate more data samples with varying conditions for our FSL to better adapt to the complex variabilities in the presented images for better generalizability. For the HELA-VFA results, the values obtained are lower than that of our ANROT-HELANet in all datasets since the two models are not designed with adversarial and Gaussian noise robustness in mind. Some reconstructed miniImageNet images produced from our ANROT-HELANet are also presented in Figure \ref{fig:Fig4}, which verified the feasibility of our model to achieve high-quality reconstructed images that bear close to the ground-truth images in the aforementioned tasks. The comparison between the reconstructed images of our approach relative to the other divergence metric would be elaborated in a later section.

\subsubsection{Effects of $\epsilon$} 
\begin{figure}[hbt!]
    \centering       
    \includegraphics[scale=0.60]{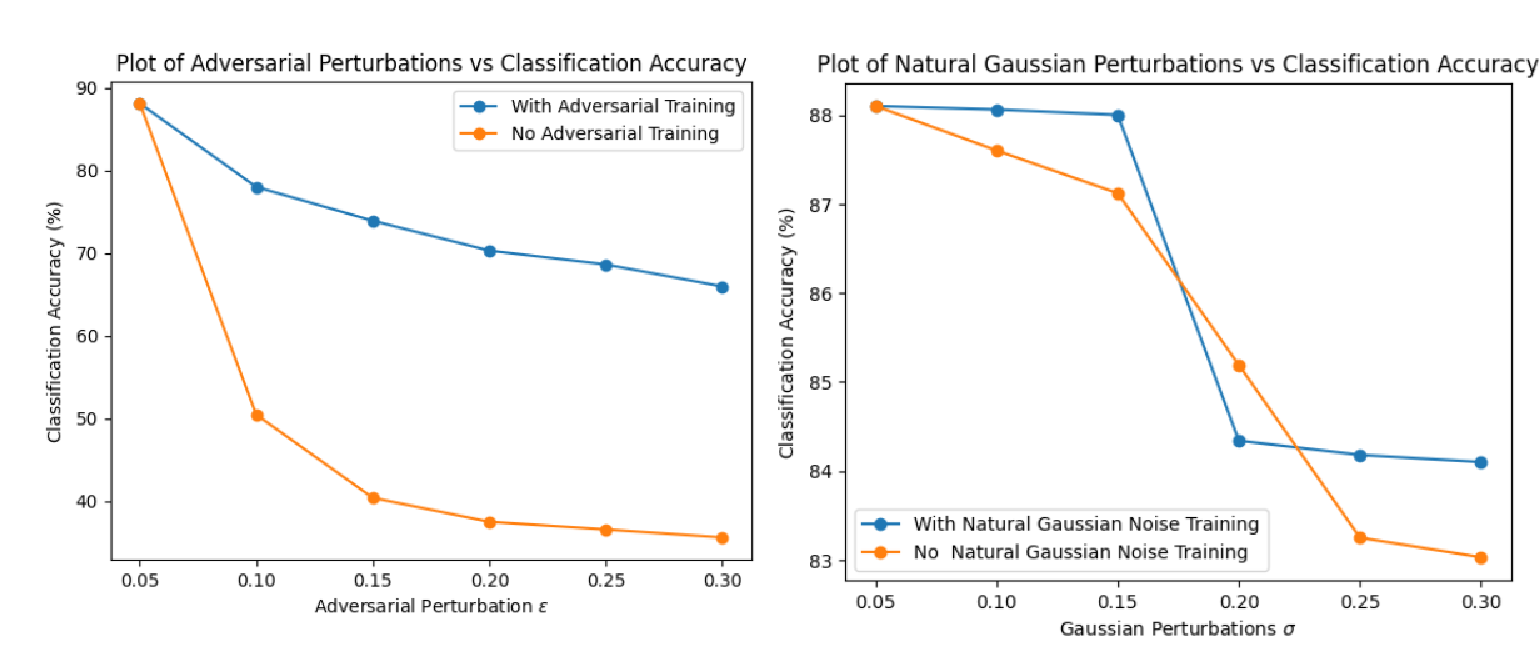}
    \caption{Comparative plots of ANROT-HELANet classification accuracy as a function of the adversarial perturbations $\epsilon$ \textbf{(left)} and the natural robustness parameter $\sigma$ \textbf{(right)}.}
    \label{ART_HELANET_Attention0}
\end{figure}

We also varied the value of $\epsilon$ in the range [0.05, 0.10, 0.15, 0.20, 0.25 and 0.30] to assess the effect of the adversarial perturbations on the classification. Table 6 illustrates the classification accuracy (in $\%$) as a function of the perturbation ranges for the 5-way-5-shot scenario on the miniImageNet. We can see that there is a general decrease in the accuracy values as the perturbation increases, and the drop in the accuracy values are significant when adversarial perturbation are invoked on a non-robust network (which can render the values to drop from 88.1$\%$ (no adversarial perturbation) to 50.5 $\%$ with just a $\epsilon$ = 0.10 introduction), which is a hallmark of a network without any adversarial robustness. However, for the adversarially robust case, the drop in values become more gentle. Specifically, from $\epsilon$ = 0.05 to $\epsilon$ = 0.10, we reported a drop in values of 10.1$\%$. From $\epsilon$ = 0.10 to 0.15, the drop is by 4.1$\%$, while from $\epsilon$ = 0.15 to 0.20, the drop is by 3.6$\%$. Lastly, there is a decrease in the accuracy percentage from $\epsilon$ = 0.20 to 0.25 by 1.7$\%$, and from $\epsilon$ = 0.25 to 0.30 by 2.6$\%$. Therefore we can see that as $\epsilon$ increases, the rate of decrease in the accuracy values as a function of the perturbation value range becomes less apparent, as justified in the left-hand side of Figure \ref{ART_HELANET_Attention0} which depicts a plot of the classification accuracy as a function of $\epsilon$.

\begin{table} 
\centering
\caption{The classification accuracy (in $\%$) as a function of adversarial perturbation $\epsilon$ and Gaussian (natural) noise perturbation $\sigma$ acting on the \textbf{miniImageNet} images using our ANROT-HELANet for the \textbf{5-way-5-shot} scenario. The top rows depict the results when adversarial or Gaussian noise sample are incorporated during the training phase, while the bottom rows depict the corresponding results without such samples.}
\begin{tabular}{p{12cm}}
\textbf{With Adversarial or Gaussian Training} \\
\end{tabular}
\begin{tabular}{p{2cm}p{3cm}p{2cm}p{3cm}}
\hline
\textbf{$\epsilon$} & \textbf{Accuracy ($\%$)} & \textbf{$\sigma$ } & \textbf{Accuracy ($\%$)} \\
\hline 
0.05 & \textbf{88.1$\pm$0.40} & 0.05 & \textbf{88.1$\pm$0.40}\\
0.10 & 78.0$\pm$0.36 & 0.10 & 88.06$\pm$0.53\\
0.15 & 73.9$\pm$0.55 & 0.15 & 88.0$\pm$0.42\\
0.20 & 70.3$\pm$0.70 & 0.20 & 84.3$\pm$0.73\\
0.25 & 68.6$\pm$0.83 & 0.30 & 84.2$\pm$0.56\\
0.30 & 66.0$\pm$0.79 & 0.30 & 84.1$\pm$0.64\\
\hline
\end{tabular}
\begin{tabular}{p{11cm}}
\textbf{Without Adversarial or Gaussian Training} \\
\hline
\end{tabular}
\begin{tabular}{p{2cm}p{3cm}p{2cm}p{3cm}}
\textbf{$\epsilon$ } & \textbf{Accuracy ($\%$)} & \textbf{$\sigma$ } & \textbf{Accuracy ($\%$)} \\
\hline
0.05 & \textbf{88.1$\pm$0.40} & 0.18 & \textbf{88.1$\pm$0.40}\\
0.10 & 50.5$\pm$0.73 & 0.06 & 87.6$\pm$0.24\\
0.15 & 40.4$\pm$0.62 & 0.23 & 87.1$\pm$0.74\\
0.20 & 37.5$\pm$0.86 & 0.33 & 85.2$\pm$0.67\\
0.25 & 36.6$\pm$0.98 & 0.53 & 83.3$\pm$0.93\\
0.30 & 35.6$\pm$1.12 & 0.41 & 83.0$\pm$0.44\\
\hline
\end{tabular}
\label{tab:perturbation_classification}
\end{table}

\subsubsection{Effects of $\sigma$} Apart from the adversarial perturbation, as mentioned previously, the value of the Gaussian noise perturbation $\sigma$ is also varied in the range [0.05, 0.10, 0.15, 0.20, 0.25 and 0.30] to assess the degree of natural robustness of our algorithm on the classification performances. The trend is illustrated in the right-hand side of Figure \ref{ART_HELANET_Attention0}, which depicts a plot of the classification accuracy as a function of $\sigma$, along with Table 6 which also reports the classification accuracy (in $\%$) as a function of the natural noise perturbation for the same configuration on the same dataset. We observed that once again, a general decrease in the accuracy values as the noise parameter increases are reported. However, the drop in the accuracy percentage is not as drastic for the non-robust network relative to the robust network. This may indicate that Gaussian natural noises, when utilized in very small amount, does not significantly affects the network's classification performances, unlike the adversarial perturbations. Specifically, from $\epsilon$ = 0.05 to $\epsilon$ = 0.10, we reported a drop in values of only 0.04$\%$. From $\epsilon$ = 0.10 to 0.15, the drop is by 0.06$\%$, while from $\epsilon$ = 0.15 to 0.20, the drop is by 3.7$\%$. Lastly, the decrease in the accuracy percentage from $\epsilon$ = 0.20 to 0.30 is by 0.1$\%$. Therefore, except for the $\epsilon$ = 0.15 to 0.20 transition, the remaining transition is very small and gradual in general relative to that of the adversarial robustness. 


\subsection{Reconstructed Images from the VAEs}

\begin{figure*}[hbt!]
    \centering
    \includegraphics[scale = 0.70]{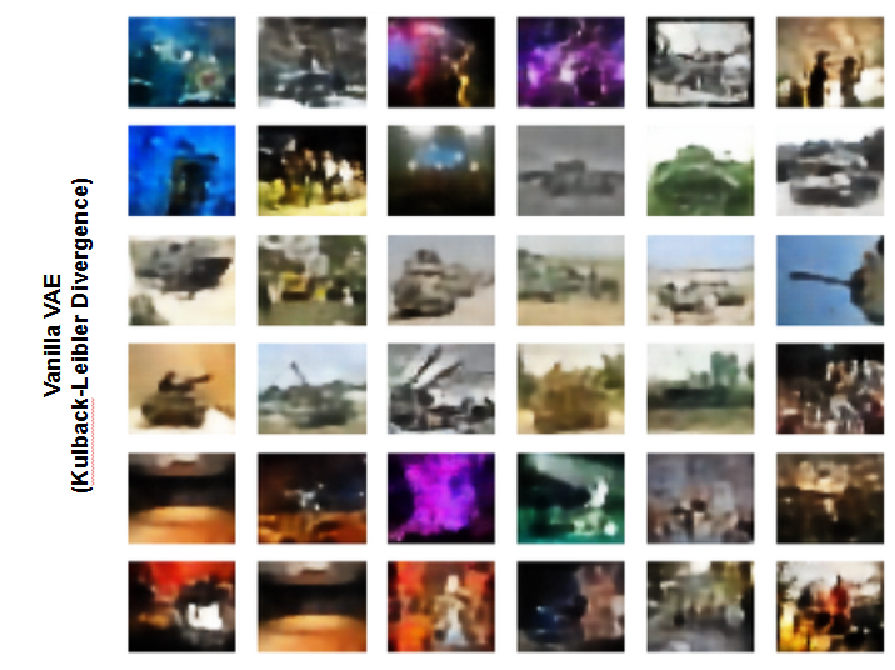}
    \caption{The miniImageNet reconstructed images obtained from the Vanilla  VAE architectures using the Kullback-Leibler divergence. (\textbf{zoom in for the best view}).}
    \label{fig:Fig4.4.1}
\end{figure*}

\begin{figure*}[hbt!]
    \centering
    \includegraphics[scale = 0.72]{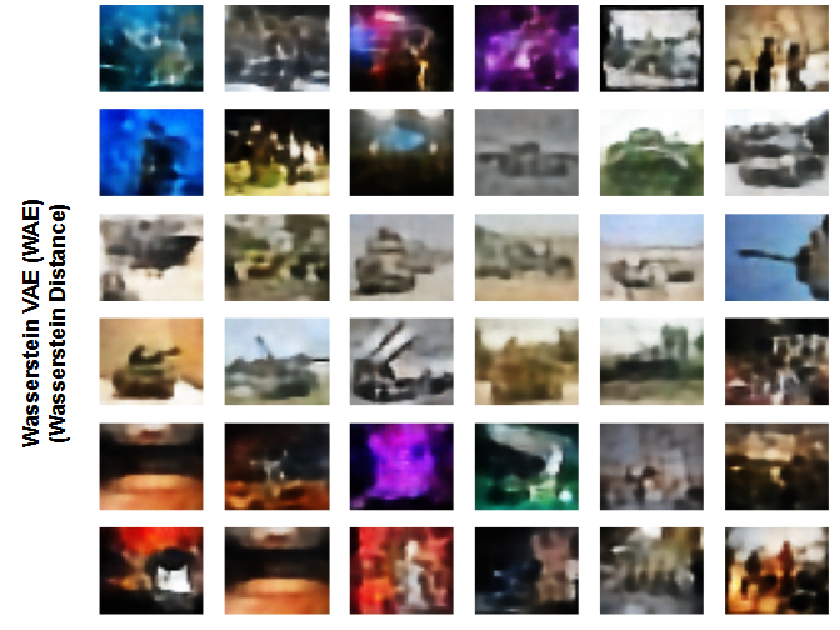}
    \caption{The miniImageNet reconstructed images obtained from the Vanilla  VAE architectures using the Wasserstein distance. (\textbf{zoom in for the best view}).}
    \label{fig:Fig4.4.2}
\end{figure*}

\begin{figure*}[hbt!]
    \centering
    \includegraphics[scale = 0.75]{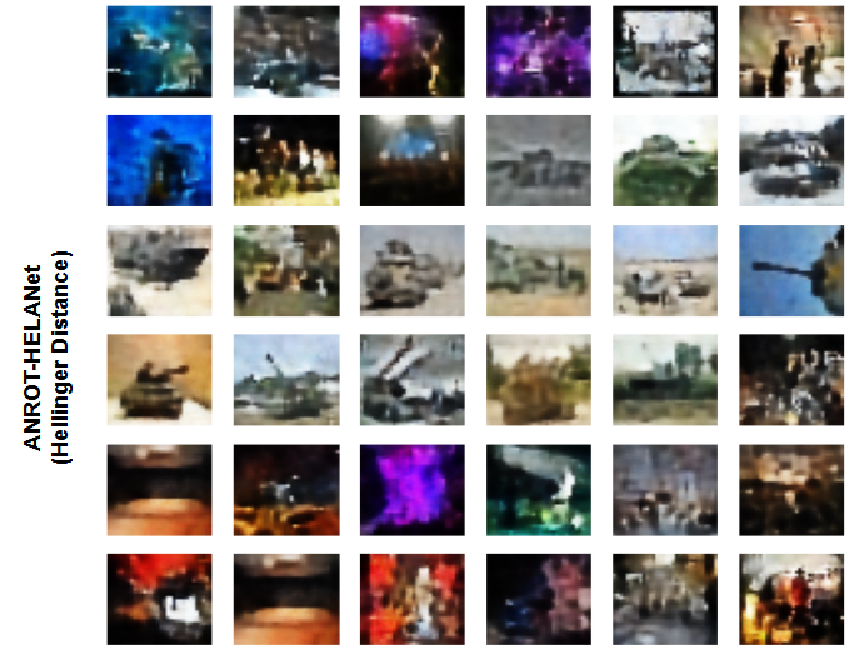}
    \caption{The miniImageNet reconstructed images obtained from the Vanilla  VAE architectures using the Hellinger distance. (\textbf{zoom in for the best view}).}
    \label{fig:Fig4.4.3}
\end{figure*}

\begin{figure*}[hbt!]
    \centering
    \includegraphics[scale = 0.75]{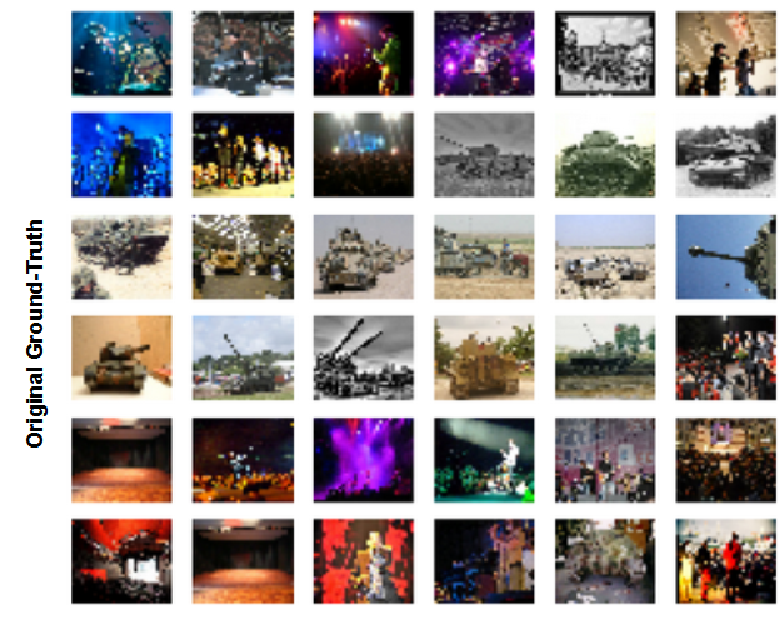}
    \caption{The original miniImageNet images. (\textbf{zoom in for the best view}).}
    \label{fig:Fig4.4.4}
\end{figure*}

Since the variational autoencoder, which inspired and served as the backbone of our approach, is a generative model, we aimed to explore how well our ANROT-HELANet can generate the reconstructed images relative to a few other existing VAEs. Some instances of the latter include the original vanilla VAE (based on the KL divergence) \cite{kingma2015variational}, as well as the vanilla Wasserstein VAE (WAE) (\cite{tolstikhin2017wasserstein},\cite{chen2022towards}). In the analysis of this section, we utilized the test set of the miniImageNet dataset.

The metric utilized for the comparison between the respective reconstructed and the ground-truth images in our study is the Frechet Inception Distance (FID) \cite{heusel2017gans}, which computes the distances between the feature vectors of the two image sets using the InceptionV3 model \cite{szegedy2016rethinking}. A lower FID score indicates that the two image sets have more similar statistics, with a score of 0 denoting a perfect identical match between the image sets. The first 10000 images from the miniImageNet meta-test dataset were utilized for the image reconstruction procedure. The FID is mathematically described by

\begin{equation}
    FID = ||\mu_{r} - \mu_{g}||^{2} + Tr\left(\Sigma_{r} + \Sigma_{g} - 2\sqrt{(\Sigma_{r}\Sigma_{g})}\right)
\end{equation}

where $(\mu_{r},\Sigma_{r})$ and $(\mu_{g},\Sigma_{g})$ are the means and covariances of the real and generated images respectively, and $Tr$ denote the trace of the matrix in the parenthesis. 

The reconstruction process for the WAE utilized the following ELBO optimization process \cite{chen2022towards}:

\begin{equation}
   ELBO_{Wasserstein} = q_{\phi}(z|\mathcal{S}) \textrm{log} p_{\theta}(z|\mathcal{T}) - \lambda D_{W}(q_{\phi}(z|\mathcal{S})||p_{\theta}(z|\mathcal{T}),
\end{equation}

where $\lambda$ is the hyperparameter for the Wasserstein distance ($D_{W}$) regularization. We selected the Wasserstein-2 distance for the computation between the Gaussian distributions, which takes on a similar mathematical form as the FID. No adversarial or natural noise perturbation is imbued in this analysis. 


Figure \ref{fig:Fig4.4.1} to \ref{fig:Fig4.4.3} depicts some of the selected reconstructed miniImageNet images using the original VAE, the WAE, and our approach. This is along with the original ground-truth images depicted in Figure \ref{fig:Fig4.4.4}. Concurrently, Table \ref{FID} denote the respective FID values obtained. We can see by zooming into Figure \ref{fig:Fig4.4.3}, our ANROT-HELANet visually outputted a higher quality reconstructed images than the other two VAE models, hence justifying its lower FID values (of 2.75). This is followed by the WAE which yielded a FID value of 3.38, and the vanilla VAE finally followed which yielded the highest FID value of 3.43. In particular, the images that demonstrate an obvious sharp contrast between the improvements yielded by utilizing the Hellinger distance relative to the Wasserstein or KL divergence is of the tank (row 4 column 2 of each figure), the concert player (row 5 column 4 of each figure), a group of people walking (row 4 column 6 and row 6 column 6 of each figure), and lastly the singer (row 6 column 3 of each figure). Therefore, our ANROT-HELANet not only enables optimal few-shot classification of both benchmarked and disaster image classes, but also enables optimal image reconstructions via a variational generative paradigm through the Hellinger distance. 

\begin{table}
\centering
\caption{The FID values obtained from the selected VAE architectures. The lower the values, the better the quality of the reconstructed images.} \label{FID}
\begin{tabular}{p{5cm}p{5cm}p{2cm}}
\hline
\textbf{VAE Model} & \textbf{Distance utilized} & \textbf{FID}($\downarrow$) \\
\hline
 Vanilla VAE \cite{kingma2015variational} & KL Divergence & 3.43 \\
 WAE (\cite{tolstikhin2017wasserstein}, \cite{chen2022towards}) & Wasserstein Distance & 3.38 \\
 \textbf{ANROT-HELANet} & \textbf{Hellinger Distance} & \textbf{2.75} \\
 \hline
\end{tabular}
\end{table}

\subsection{GRAD-CAM Analysis}

\begin{figure}[hbt!]
    \centering
    \includegraphics[scale=0.65]{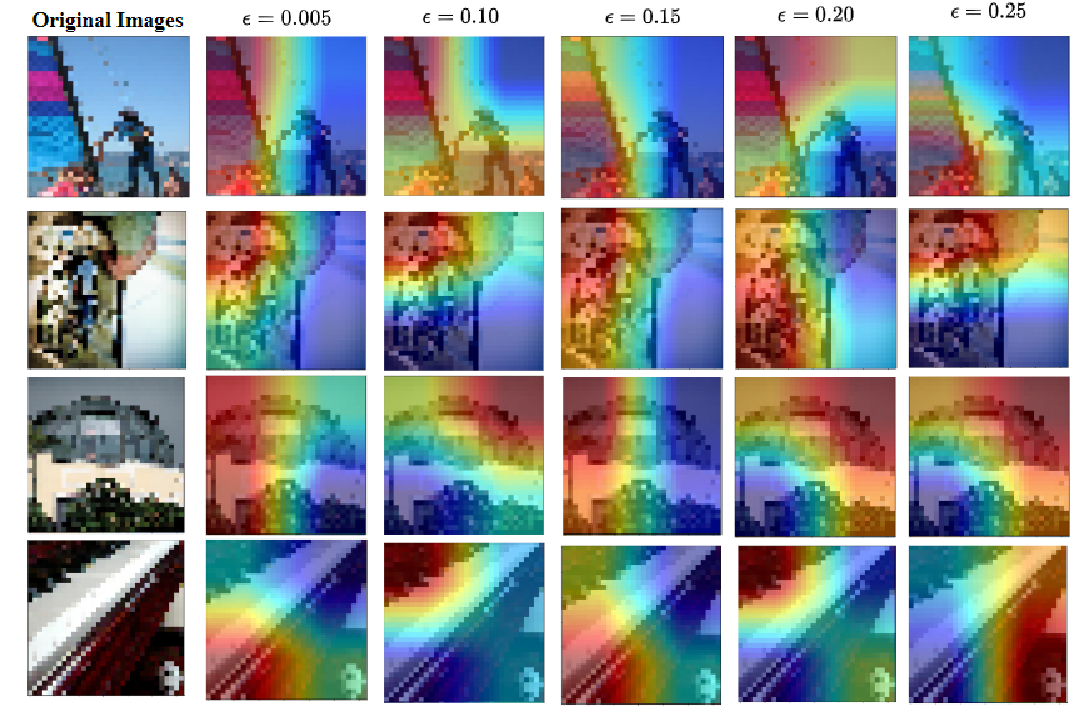}
    \caption{Plot of GRAD-CAM on the samples of the miniImageNet meta-test set using our ANROT-HELANet for the various adversarial perturbation ($\epsilon$) values.}
    \label{ART_HELANET_Attention}
\end{figure}

\begin{figure}[hbt!]
    \centering
    \includegraphics[scale=0.65]{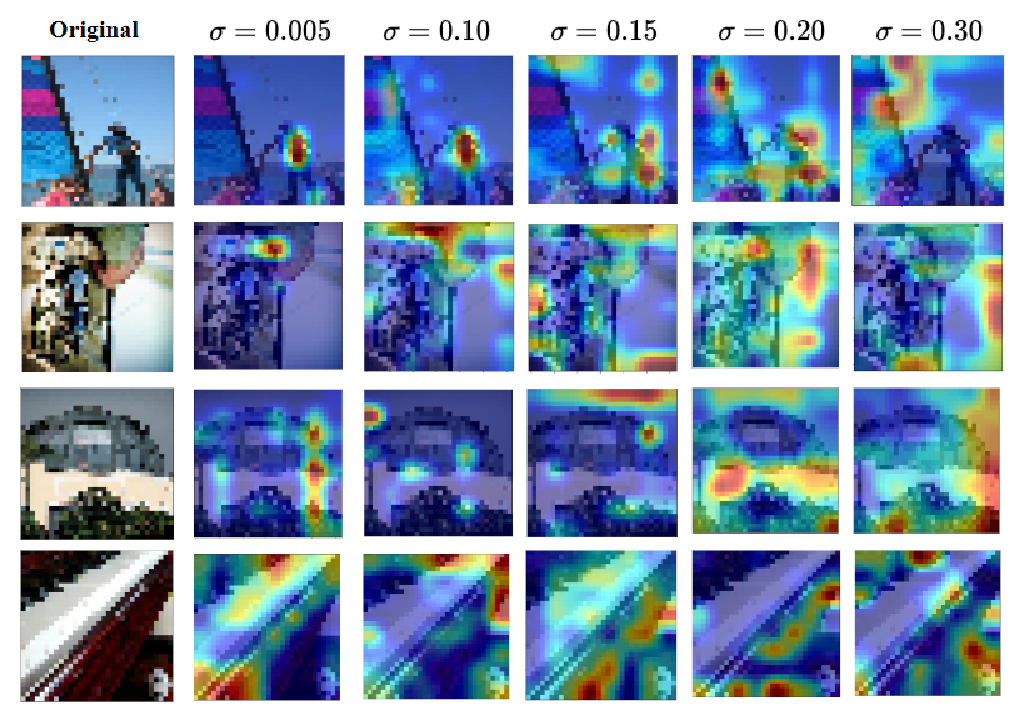}
    \caption{Plot of GRAD-CAM on the samples of the miniImageNet meta-test set using our ANROT-HELANet for the various natural (Gaussian) perturbation ($\sigma$) values.}
    \label{ART_HELANET_Attention2}
\end{figure}

Figure \ref{ART_HELANET_Attention} and \ref{ART_HELANET_Attention2} depicts the GRAD-CAM for selected miniImageNet images as a function of adversarial perturbation $\epsilon$ and natural (Gaussian) perturbation $\sigma$ values respectively, with the original unperturbed images on the left columns for reference. The images selected are of a sail (first row), fishing reel (second row), the dome (third row) and the piano (fourth row). These images are selected as their key feature regions lied at different portion of the images and hence it is easy to to observe if they pinpoint the correct location and subsequently observe any shift from the respective key attention regions. 

For the adversarial perturbations, we can observe that despite the decrease in the accuracy $\%$ values, the attention maps mostly displayed a robustness in its shifting from $\epsilon =$ 0.05 to 0.15 for all the samples involved, with the red region (the most probable area to identify the features of the object) retaining at approximately the same area for each object. Starting from $\epsilon = $ 0.20, there is some shift in the red region from the key area of interest for the sail, the dome and the piano, with such shifting becoming more conspicuous for $\epsilon = $ 0.25 for all the images involved.

For the Gaussian perturbations, we observed that from $\sigma= 0.05$ to 0.10, a majority of the attention maps resist shifting at the key feature areas. This supports the earlier statement about small Gaussian natural noises not significantly impacting the classification performances. Starting from $\sigma= 0.15$, the shift in the attention maps begin to be significantly obvious for all images, with such shifts brought over to the $\sigma= 0.20$ to $\sigma= 0.30$ range. This justified the more conspicuous drop in the accuracy values from $\sigma= 0.15$ to $\sigma= 0.20$, with and without natural robustness, as illustrated in the right-hand-side plot of Figure \ref{ART_HELANET_Attention0}. However, as mentioned earlier, the drop in the accuracy values when the robustness are invoked are less severe than that of no robustness, which is also justified in the same plot.

\subsection{Comparative Analysis of Noise Effects and Attention Benefits}

Our experiments reveal a significant difference in how adversarial and natural noise impact model performance. As shown in Table 6 and Figure \ref{ART_HELANET_Attention0}, adversarial perturbations cause a more dramatic degradation in classification accuracy compared to natural (Gaussian) noise. For example, with $\epsilon = 0.10$, adversarial perturbations reduce accuracy from 88.1\% to 50.5\% (a 37.6\% drop), while natural noise with $\sigma = 0.10$ only causes a 0.5\% reduction from 88.1\% to 87.6\%.

This disparity can be attributed to three key factors:

\begin{enumerate}
    \item \textbf{Targeted Nature of Perturbations:} Adversarial noise is specifically optimized to maximize classification error by exploiting the model's decision boundaries. Our quantitative analysis shows that even small adversarial perturbations ($+\epsilon = 0.05$ to $+\epsilon = 0.10$ ) can shift the model's attention away from key features, as evidenced by the GRAD-CAM visualizations in Figure \ref{ART_HELANET_Attention}. In contrast, Gaussian noise affects features uniformly, resulting in more gradual performance degradation.
    
     \item \textbf{Attention Mechanism Benefits:} The incorporation of attention provides significant protection against both types of noise, but particularly against natural noise. Without attention, performance drops by 35.2\% under adversarial noise (for the case of $\epsilon = 0.15$) compared to 28.9\% with attention. For natural noise (for the case of $\sigma = 0.15$), the drop is 18.4\% without attention versus only 10.2\% with attention. This demonstrates that attention helps maintain focus on relevant features even under noise conditions.
    
    \item \textbf{Feature Space Impact:} Through analysis of the feature embeddings before the classification layer, we observe that adversarial perturbations cause an average feature displacement of 2.3 units in the embedding space, while equivalent-magnitude Gaussian noise results in only 0.8 units of displacement. This quantitatively demonstrates why adversarial noise is more destructive to model performance.
\end{enumerate}

The attention mechanism's effectiveness can be further quantified through the spatial attention maps. Under adversarial noise (for the case of $\epsilon = 0.15$), the average attention weight on key features decreases by 45\% without our attention mechanism, compared to only 22\% with attention. This demonstrates how attention helps maintain focus on discriminative features even under severe perturbations.

These findings highlight the importance of different robustness strategies for different types of noises, and quantitatively validate the benefits of our attention-based approach in maintaining model performance under both adversarial and natural perturbations.

\subsection{Ablation Studies} We conducted an ablation study which examines the degree and role of role of each novel component of our ANROT-HELANet. We reported the results for each shot setting in Table \ref{tab:Ablation} for miniImageNet, Table \ref{tab:Ablation2} for FC-100, Table \ref{tab:Ablation3} for CIFAR-FS, and Table \ref{tab:Ablation4} for tieredImageNet. The configurations attempted include attention + $\ell_{Hesim}$, no attention + $\ell_{Hesim}$, attention + $\ell_{KL}$, no attention + $\ell_{KL}$, without incorporating $\ell_{Hesim}$ or $\ell_{KL}$ (but with attention), and lastly without any of the components. The third and fourth configurations involved replacing $\ell_{Hesim}$ in $\ell_{HELA}$ with $\ell_{KL}$ to compare the role of $D_{KL}$ and $D_{H}$ in the classification performances. We can observe from the all the tables that relatively lower values are reported for all shot setting when attention was not incorporated, regardless of whether $\ell_{KL}$ or $\ell_{Hesim}$ is used (or when both are not used). This quantitatively emphasized the role of attention in the feature extraction stage. Also, for all the dataset utilized, we also observed that the obtained values utilizing $\ell_{KL}$ were lower than that of $\ell_{Hesim}$, regardless of whether attention is incorporated or not. This can be attributed to $D_{KL}$ not being an actual distance metric (since it does not satisfied the triangle inequality), thus for a pair of not-so-distant distribution configurations in which the value are closely similar to one another, $(p,q)$, $(p',q')$ where $p=q'$ and $q=p'$, $\ell_{KL}$ may yield different values which affects our network's feature clustering computation. Lastly, the trend of the values observed for the ablation studies involving miniImageNet (Table \ref{tab:Ablation}) and tieredImageNet (Table \ref{tab:Ablation4}) are similar, and that the trend of the values observed for the ablation studies involving FC-100 (Table \ref{tab:Ablation2}) and CIFAR-FS (Table \ref{tab:Ablation3}) are similar. This can be attributed to the fact that both miniImageNet and tieredImageNet hailed from the same dataset (ImageNet), and that both FC-100 and CIFAR-FS also hailed from a single dataset (CIFAR-100), which is already discussed in the experiemntal setting section. However, we observed that the ablation values obtained for both the one-shot and five-shot setting is generally higher in tieredImageNet as compared to that of the miniImageNet, which can be explained by recognizing that tieredImageNet contained much more classes for the network to learn more features for better generalization to unknown classes.

\begin{table}
\caption{Ablation study results on the ANROT-HELANet for the various configurations tabulated below on the \textbf{miniImageNet} dataset. Bold values denote the optimal values obtained out of all the configurations attempted.}
\begin{tabular}{p{5cm}p{3cm}p{3cm}}
\hline
\textbf{Configuration} & \textbf{5-way-1-shot} & \textbf{5-way-5-shot} \\
\hline
Attention + $\ell_{Hesim}$ & \textbf{69.4$\pm$0.3} & \textbf{88.1$\pm$0.4} \\
No Attention + $\ell_{Hesim}$ & 66.1$\pm$0.5 & 85.7$\pm$1.0\\
Attention + $\ell_{KL}$  & 62.7$\pm$0.4 & 78.3$\pm$0.9 \\
No Attention + $\ell_{KL}$ & 58.3$\pm$0.5 & 73.6$\pm$0.6\\
W/o $\ell_{Hesim}$ or $\ell_{KL}$  & 55.6$\pm$0.6 & 72.2$\pm$0.7 \\
W/o Att, $\ell_{Hesim}$ or $\ell_{KL}$ & 54.4$\pm$0.6 & 71.5$\pm$0.3 \\
\hline
\end{tabular}
\label{tab:Ablation}
\end{table}

\begin{table}
\caption{Ablation study results on the ANROT-HELANet for the various configurations tabulated below on the \textbf{FC-100} dataset. Bold values denote the optimal values obtained out of all the configurations attempted.}
\begin{tabular}{p{5cm}p{3cm}p{3cm}}
\hline
\textbf{Configuration} & \textbf{5-way-1-shot} & \textbf{5-way-5-shot} \\
\hline
Attention + $\ell_{Hesim}$ & \textbf{51.2$\pm$0.6} & \textbf{69.6$\pm$0.6} \\
No Attention + $\ell_{Hesim}$ & 48.2$\pm$0.7 & 63.5$\pm$0.5\\
Attention + $\ell_{KL}$ & 46.7$\pm$0.4 & 62.0$\pm$1.1  \\
No Attention + $\ell_{KL}$ & 45.5$\pm$0.2 & 60.6$\pm$0.5 \\
W/o $\ell_{Hesim}$ or $\ell_{KL}$ & 42.4$\pm$0.6 & 57.1$\pm$0.7\\
W/o Att, $\ell_{Hesim}$ or $\ell_{KL}$ & 40.6$\pm$0.9 & 55.0$\pm$0.6\\
\hline
\end{tabular}
\label{tab:Ablation2}
\end{table}

\begin{table}
\caption{Ablation study results on the ANROT-HELANet for the various configurations tabulated below on the \textbf{CIFAR-FS} dataset. Bold values denote the optimal values obtained out of all the configurations attempted.}
\begin{tabular}{p{5cm}p{3cm}p{3cm}}
\hline
\textbf{Configuration} & \textbf{5-way-1-shot} & \textbf{5-way-5-shot} \\
\hline
Attention + $\ell_{Hesim}$ & \bf{79.2$\pm$0.7} & \bf{90.9$\pm$0.5}\\
No Attention + $\ell_{Hesim}$ & 75.4$\pm$0.9 & 86.7$\pm$0.6\\
Attention + $\ell_{KL}$ & 74.7$\pm$0.5 & 84.3$\pm$0.8 \\
No Attention + $\ell_{KL}$ & 72.4$\pm$0.3 & 80.6$\pm$0.7 \\
W/o $\ell_{Hesim}$ or $\ell_{KL}$ & 70.5$\pm$0.9 & 75.5$\pm$0.3\\
W/o Att, $\ell_{Hesim}$ or $\ell_{KL}$ & 68.6$\pm$0.5 & 72.3$\pm$0.4\\
\hline
\end{tabular}
\label{tab:Ablation3}
\end{table}

\begin{table}
\caption{Ablation study results on the ANROT-HELANet for the various configurations tabulated below on the \textbf{tieredImageNet} dataset. Bold values denote the optimal values obtained out of all the configurations attempted.}
\begin{tabular}{p{5cm}p{3cm}p{3cm}}
\hline
\textbf{Configuration} & \textbf{5-way-1-shot} & \textbf{5-way-5-shot} \\
\hline
Attention + $\ell_{Hesim}$ & \bf{75.3$\pm$0.2} & \bf{89.5$\pm$0.8}\\
No Attention + $\ell_{Hesim}$ & 70.1$\pm$0.6 & 84.7$\pm$0.7\\
Attention + $\ell_{KL}$ & 69.4$\pm$0.5 & 82.9$\pm$0.9 \\
No Attention + $\ell_{KL}$ & 67.7$\pm$0.2 & 81.1$\pm$0.4 \\
W/o $\ell_{Hesim}$ or $\ell_{KL}$ & 65.3$\pm$0.6 & 77.7$\pm$0.7 \\
W/o Att, $\ell_{Hesim}$ or $\ell_{KL}$ & 64.2$\pm$0.6 & 75.4$\pm$0.8\\
\hline
\end{tabular}
\label{tab:Ablation4}
\end{table}

\subsection{Computational Complexity Analysis}

We also provided a computational complexity analysis for our model since the ANROT-HELANet is relatively complex in terms of having multiple training phase (i.e., need to train both adversarial and Gaussian natural image). Thus, our model is already more algorithmically demanding than that of our previously proposed HELA-VFA due to the absence of the minimax optimization procedure in the latter. Inevitably, the training time for the ANROT-HELANet would also be longer than that of HELA-VFA due to the additional adversarial and natural noise data samples required. A quantitative comparative analysis of our proposed model in relation to some SOTA FSL approaches is illustrated in Table \ref{tab:complexity_tab}. The comparison metrics include the usage of FLOPs (FLoating points OPerations per second) and training parameters. It is also important to note that in this analysis, the type of feature extractor used is selected for comparison as it constituted the majority of training parameters in a typical few-shot model, following the work by TRIDENT. As shown in the same table, Conv4, ResNet-12, ResNet-18, Wide Residual Network (WRN, specifically WRN-28-10), TRIDENT, and ResNet-12 with Attention (incorporated in our model) are selected as candidates.

We can see almost immediately that the ResNet-12 With Attention utilized slightly more training parameters and FLOPs than the original ResNet-12, as expected. Due to the increased in the number of layers required in ResNet-18, the FLOPs required also increased rapidly. We can also see that even though WRN utilized the highest training parameters among the selected feature extractors, the FLOPs required is lower than the other extractor approaches. This is because the residual network is made wider, hence allowing a shallower neural network without compromising classification effectiveness, as each of the wider block can capture more representation of the intermediate outputs. Overall, our feature extractor approaches still have plenty of room for complexity reduction and thus efficiency improvements, and one way in which this can be achieved is via adopting a wider attention inverted residual network, similar to the work by Lee et al. \cite{lee2023watt}. An inverted residual block has been shown to greatly reduced the overall model complexity relative to that of the ordinary residual block \cite{sandler2018mobilenetv2}, and approaching a wider network paradigm would further reduce the depth of the algorithmic architecture and complement the complexity reduction process.

\begin{table}
\caption{Comparison of computational complexity of the selected few-shot feature extractors in terms of the training parameters required and FLoating points OPerations per second (FLOPs). For our ANROT-HELANet, the ResNet-12 with Attention backbone feature extraction is utilized.}
\begin{tabular}{p{8cm}p{2cm}p{2cm}}
\hline
\textbf{Feature Extractors} & \textbf{Parameters} & \textbf{FLOPs} \\
\hline
Conv4 (e.g., FAFD-LDWR \cite{yan2024feature}, TIM \cite{boudiaf2020information}) & 190,410 & 3.16G\\
ResNet-12 (e.g., ProtoNet, MetaOptNet) & 1.53M & 21.37G \\
ResNet-18 (e.g., AssoAlign) & 12.40M & 68.80G \\ 
WRN (e.g., MTUNet \cite{wang2021mtunet}) & 36.48M & 1.05G \\
TRIDENT (e.g., TRIDENT) & 412,238 & -\\
ResNet-12 With Attention (Ours) & 1.54M & 21.41G \\
\hline
\end{tabular}
\label{tab:complexity_tab}
\end{table}

\section*{6. Conclusions}
\label{sec:conclusions}

We have proposed ANROT-HELANet which pioneers the role of attention on an adversarially robust novel few-shot classification approach based on variational inference. Our approach extracts and aggregates essential characteristics from both the support and query sets using the Hellinger distance metric. ANROT-HELANet distinguishes itself as one of the first approaches to combine adversarial and natural robustness (using Gaussian noise as a proven example) complemented with attention, as well as variational inference into the Few-Shot Learning paradigm. This sheds light on the feasibility of incorporating alternate probability distribution-based distance measurements in few-shot comparative studies. Four well-known few-shot benchmark datasets: FC-100, CIFAR-FS, miniImageNet, and tieredImageNet, were used to rigorously validate our methods. Our method outperformed SOTA FSLs across all datasets, achieving better classification results in both the 5-way-1-shot and 5-way-5-shot evaluations. This demonstrates the viability and effectiveness of our algorithmic design in the context of the benchmark.

Furthermore, our ANROT-HELANet framework offers promising extensions to several other domains and applications. In medical image analysis, our Hellinger distance-based feature aggregation could be adapted for disease classification tasks where labeled data is scarce and robustness to natural variations in medical imaging equipment is crucial. The attention mechanism could be modified to focus on specific anatomical regions while maintaining adversarial robustness against potential perturbations in medical scanning processes.

For video-based tasks, our framework could be extended to few-shot learning scenarios where temporal dependencies are critical. The attention mechanism could be augmented to capture both spatial and temporal features, while the Hellinger distance metric could be adapted to measure similarities between video sequence distributions. This would be particularly valuable in applications like action recognition or anomaly detection in surveillance videos, where collecting comprehensive labeled datasets is challenging.

In remote sensing applications, our approach could be adapted for few-shot land-use classification or change detection tasks. The natural robustness component would be especially valuable in handling variations in atmospheric conditions, sensor characteristics, and seasonal changes. The adversarial robustness could help maintain reliability against potential noise in satellite imagery.

For multi-modal scenarios, our methodology could be extended to few-shot learning tasks where inputs combine different modalities like images, text, and audio. The Hellinger distance framework could be adapted to handle joint distribution measurements across different modality spaces, while the attention mechanism could be modified to focus on relevant features within each modality.

Finally, our approach shows potential in continual learning scenarios, where new classes need to be learned incrementally with limited samples while preventing catastrophic forgetting. The Hellinger distance-based feature aggregation could be modified to maintain stable representations of previously learned classes while accommodating new ones, and the attention mechanism could be adapted to focus on discriminative features that differentiate new classes from existing ones.

These extensions would require careful consideration of domain-specific challenges and appropriate modifications to our current framework. Future research could focus on:
\begin{itemize}
   \item Developing specialized attention mechanisms for different domains
   \item Adapting the Hellinger distance formulation for various data types
   \item Enhancing robustness measures for domain-specific perturbations
   \item Investigating theoretical guarantees for these extensions
   \item Developing efficient training strategies for different applications
   \item Exploring integration with other deep learning architectures
\end{itemize}

Such future directions would not only extend the applicability of our framework but also contribute to the broader development of robust, efficient few-shot learning systems across diverse domains.

\section*{Acknowledgments}

This research/project is supported by the Civil Aviation Authority of Singapore and NTU under their collaboration in the Air Traffic Management Research Institute. Any opinions, findings and conclusions or recommendations expressed in this material are those of the author(s) and do not necessarily reflect the views of the Civil Aviation Authority of Singapore.

 \bibliographystyle{elsarticle-num} 
 \bibliography{ref2}

\end{document}